\documentclass[sigconf]{acmart}

\copyrightyear{2023}
\acmYear{2023}
\setcopyright{rightsretained}
\acmConference[CIKM '23]{Proceedings of the 32nd ACM International Conference on Information and Knowledge Management}{October 21--25, 2023}{Birmingham, United Kingdom}
\acmBooktitle{Proceedings of the 32nd ACM International Conference on Information and Knowledge Management (CIKM '23), October 21--25, 2023, Birmingham, United Kingdom}\acmDOI{10.1145/3583780.3614995}
\acmISBN{979-8-4007-0124-5/23/10}

\makeatletter
\gdef\@copyrightpermission{
  \begin{minipage}{0.3\columnwidth}
   \href{https://creativecommons.org/licenses/by/4.0/}{\includegraphics[width=0.90\textwidth]{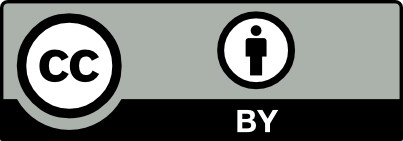}}
  \end{minipage}\hfill
  \begin{minipage}{0.7\columnwidth}
   \href{https://creativecommons.org/licenses/by/4.0/}{This work is licensed under a Creative Commons Attribution International 4.0 License.}
  \end{minipage}
  \vspace{5pt}
}
\makeatother

\usepackage[T1]{fontenc}
\usepackage{amsfonts}
\usepackage{multirow}
\usepackage{balance}
\usepackage{bbm}
\usepackage{graphicx}
\usepackage[linesnumbered,ruled,vlined]{algorithm2e}
\usepackage{enumitem}
\usepackage[flushleft]{threeparttable}
\usepackage{subcaption}
\SetCommentSty{mycommfont}
\newcommand{\Pa}[1]{\mathsf{PA}_{#1}}
\newcommand{\pa}[1]

\newcommand{\x}{\mathbf{x}}

\newcommand{\z}{\mathbf{z}}

\usepackage{mathtools}

\usepackage{amsmath}
\allowdisplaybreaks

\DeclareMathOperator*{\argmin}{arg\,min}
\newtheorem{definition}{Definition}
\usepackage{color}

\author{Xiao Han}
\orcid{0000-0002-1953-8658}
\affiliation{%
  \institution{Utah State University}
  \city{Logan}
  \state{UT}
  \postcode{84322}
  \country{USA}
}
\email{xiao.han@usu.edu}

\author{Lu Zhang}
\orcid{0000-0002-8972-8799}
\affiliation{%
  \institution{University of Arkansas}
  \city{Fayetteville}
  \state{AR}
  \postcode{72701}
  \country{USA}
}
\email{lz006@uark.edu}

\author{Yongkai Wu}
\orcid{0000-0002-7313-9439}
\affiliation{%
  \institution{Clemson University}
  \city{Clemson}
  \state{SC}
  \postcode{29634}
  \country{USA}
}
\email{yongkaw@clemson.edu}

\author{Shuhan Yuan}
\orcid{0000-0001-6816-419X}
\affiliation{%
  \institution{Utah State University}
  \city{Logan}
  \state{UT}
  \postcode{84322}
  \country{USA}
}
\email{Shuhan.Yuan@usu.edu}

\title{On Root Cause Localization and Anomaly Mitigation through Causal Inference}

\begin{document}

\begin{abstract}
Due to a wide spectrum of applications in the real world, such as security, financial surveillance, and health risk,  various deep anomaly detection models have been proposed and achieved state-of-the-art performance. However, besides being effective, in practice, the practitioners would further like to know what causes the abnormal outcome and how to further fix it. In this work, we propose RootCLAM, which aims to achieve Root Cause Localization and Anomaly Mitigation from a causal perspective. Especially, we formulate anomalies caused by external interventions on the normal causal mechanism and aim to locate the abnormal features with external interventions as root causes. After that, we further propose an anomaly mitigation approach that aims to recommend mitigation actions on abnormal features to revert the abnormal outcomes such that the counterfactuals guided by the causal mechanism are normal. Experiments on three datasets show that our approach can locate the root causes and further flip the abnormal labels.

\end{abstract}




\maketitle

\section{Introduction}
Deep anomaly detection models have been used to automatically detect a variety of anomalies, such as bank fraud detection. As many anomaly detection tasks are high-stakes decision-making tasks, there is a growing demand for the transparency of the detection results, especially, for the outcomes as anomalies \cite{panjei2022survey}. For example, if a credit card transaction is declined by an automated decision-making algorithm due to the potential fraudulent features of this transaction, the user would like to know which features lead to the transaction decline and how to avoid such a situation in the future. 

To answer the question of which features lead to abnormal outcomes, several interpretable anomaly detection approaches are proposed based on the idea of feature attributions \cite{kauffmannExplainingAnomaliesDeep2020,sippleInterpretableMultidimensionalMultimodal2020,liznerskiExplainableDeepOneClass2021}. Although feature attribution-based approaches can highlight the abnormal features, they ignore the dependencies between different features, whereas some abnormal features may be caused by other upstream abnormal features. For example, if a loan application is declined, a feature attribution-based approach may highlight the low income and low savings as abnormal features. However, the actual situation may be that low savings are caused by low income, and low income is the root cause of the loan application decline. Identifying the root cause of the anomaly can provide insights into the anomaly as well as efficient actions to fix the anomaly. 

In this paper, we study the problem of anomaly mitigation facilitated by the root cause localization. We propose a framework named Root Cause Localization and Anomaly Mitigation (RootCLAM). The framework consists of two phases. In the first phase, we attempt to identify and localize the features that are the root cause of the anomaly for each abnormal instance. Then, in the second phase, we answer the question of how to fix the abnormal outcome by finding the algorithmic recourse \cite{datta2022framing} on the abnormal outcome. Traditional algorithmic recourse may perform actions on any feature in order to improve or flip the outcome. However, in the context of anomaly mitigation, it is more natural to perform recourse actions on the root cause features as not all features are equally important for mitigation. Thus, our framework aims to find the algorithmic recourse by only using root cause features.

Developing RootCLAM faces several challenges. First, despite several root cause analysis approaches proposed for anomalies in time series data \cite{assaad2023root,yang2023causal,meng2020localizing,budhathoki2022causal}, the research on the root cause analysis of the tabular data is still limited, especially in the context of anomaly detection. Second, to perform appropriate recourse actions on root cause features to change the outcome, one needs to quantitatively analyze the causal connection between these actions and the outcome \cite{assaadRootCauseIdentification2023,yangCausalApproachDetecting2022}. Last but not least, algorithmic recourse is known as providing a counterfactual interpretation of the outcome. However, existing counterfactual inference techniques \cite{han2023achieving,ng2019graph} usually assume that the causal connections between features can be described by linear equations, which may not be realistic in practical situations.

To address these challenges, we first assume that the data generation is governed by a Structural Causal Model (SCM) \cite{pearlCausality2009}, and treat the root cause as external interventions on specific features. As a result, the root cause localization is to identify features that are impacted by the external
intervention. Then, we formulate the algorithmic recourse for anomaly mitigation as soft interventions \cite{correa2020calculus} in order to represent the causal effect of recourse actions on the outcome as a differentiable expression. Based on that, we develop a continuous optimization-based iterative algorithm that follows the causal graph topological order to compute the actions such that the outcome will be flipped to normal by performing the actions. In addition, we leverage the causal graph autoencoder to conduct counterfactual inference. In particular, we adopt the Variational Causal Graph Autoencoder (VACA) \cite{sanchez2021vaca} which can deal with non-linear SCMs by leveraging graph neural networks. Finally, anomaly mitigation is achieved as the outcome of the algorithmic recourse based on root cause features.

For empirical evaluation, we conduct experiments on several semi-synthetic and real-world datasets. The results show that our method can produce the largest flipping ratio regarding the anomaly detection outcomes while requiring the minimum perturbation compared with the baseline methods.

\section{Preliminary}

\subsection{Structural Causal Model (SCM)}
We adopt Pearl's Structural Causal Model (SCM) \cite{pearlCausality2009} as the prime methodology for computing counterfactuals.
Throughout this paper, we use the upper/lower case alphabet to represent features/values.

\begin{definition}
	An SCM is a triple $\mathcal{M}= \{U,V,F\}$ where 
	\begin{itemize}[leftmargin=*]
	    \item[ 1)] $U$ is a set of exogenous variables that are determined by factors outside the model. A joint probability distribution $P(u)$ is defined over the features in $U$.
	    \item[ 2)] $V$ is a set of endogenous variables/features that are determined by variables in $U\cup V$.
	    \item[ 3)] $F$ is a set of functions $\{f_1,\ldots,f_n\}$; for each $X_i \in V $, a corresponding function $f_{i}$ is a mapping from $U \cup (V\setminus \{X_i\})$ to $X_i$, %i.e., $x_{i}=f_{i}(x_{\Pa{i}},u_{i})$, 
        where a set of features $X_{\Pa{i}}\subseteq V\backslash \{X_{i}\}$ are called the parents of $X_{i}$.%, and $U_{i}\subseteq U$.
	\end{itemize}
\end{definition}

An SCM is often illustrated by a causal graph $\mathcal{G}$ where each observed variable is represented by a node, and the causal relationships are represented by directed edges $\rightarrow$. 

Inferring causal effects in the SCM is facilitated by the intervention. The hard intervention forces some variable $X\in V$ to take a certain value $x$. 
For an SCM $\mathcal{M}$, intervention $do(X=x')$ is equivalent to replacing original function in $F$ with $X = x'$. 
The soft intervention, on the other hand, forces some variables to take a certain functional relationship in responding to some other
variables \cite{correa2020calculus}. The soft intervention substitutes equation $x = f(x_{\Pa{}},u)$ with a new equation.
After the intervention, the distributions of all features that are the descendants of $X$ may be changed, called the interventional distributions.

\subsection{Counterfactuals}
Counterfactuals are about answering questions such as for two features $X,Y\in V$, whether $Y$ would be $y$ had $X$ been $x'$ given that $X$ is equal to $x$ in the factual instance. Symbolically we denote this counterfactual instance as $x_{do(X=x')}|x$.
The counterfactual question involves two worlds, the factual world and the counterfactual world, and cannot be answered directly by the do-operator. 
When the complete knowledge of the SCM is known, the counterfactual can be computed by the Abduction-Action-Prediction process \cite{pearlCausality2009}:
\begin{itemize}[leftmargin=*]
    \item[ 1)] Abduction: Beliefs about the world are updated by taking into account all evidence given in the context. Formally, update the probability $P(u)$ to $P(u|e)$.
    \item[ 2)] Action: Perform do intervention, $do(X=x')$, to reflect the counterfactual assumption, and a new causal model is created by interventions $\mathcal{M}' = \mathcal{M}_{do(X=x')}$.
    \item[ 3)] Prediction: Counterfactual reasoning occurs over the new model $\mathcal{M}'$ using updated knowledge $P(u|e)$.
\end{itemize}

\subsection{Causal Graph Autoencoder}
A causal graph autoencoder is a type of deep learning model that aims to learn a latent representation of the data that captures the underlying causal relationships among variables given a causal graph.
In this paper, we adopt the Variational Causal Graph Autoencoder (VACA) \cite{sanchez2021vaca} which can accurately approximate the interventional and counterfactual distributions on diverse SCMs and can deal with non-linear causal relationships. The VACA consists of an adjacency matrix $A$ of the causal graph, a decoder $p_{\zeta}(\mathbf{x}|\mathbf{z},A)$ which is a graph neural network (GNN) that takes as input a set of latent variables $\mathbf{z}$ and the matrix $A$ and outputs the likelihood of $\x$, and an encoder $q_{\xi}(\mathbf{z}|\mathbf{x},A)$ which is another GNN that takes $\mathbf{x}$ and $A$ as input and outputs the latent variables of $\z$. The VACA is trained to fit the observational distribution.

To compute the counterfactual instance of a factual instance $\x$ under the hard intervention $do(X_i = x')$, the VACA first computes the distribution of $\z$ by feeding the factual instance $\x$ and $A$ into encoder $q_{\xi}(\mathbf{z}|\mathbf{x},A)$. Then, the VACA constructs the intervened instance $\bar{\x}$ by replacing the value of $x_i$ in the factual instance $\x$ with the intervened value $x'$, as well as the intervened matrix $\bar{A}$ by removing all incoming edges of node $X_i$ in the causal graph. The VACA feeds $\bar{\x}$ and $\bar{A}$ into encoder $q_{\xi}(\mathbf{z}|\mathbf{x},A)$ to compute the intervened distribution of the latent variables, denoted by $\bar{\z}$. Next, the VACA removes the latent variable in $\z$ that corresponds to $x_i$, i.e., $z_i$, and replaces it with $\bar{z}_i$ in $\bar{\z}$ to obtain a new vector $\tilde{\z}$. This step is to perform the intervention in the hidden space that is equivalent to performing the intervention in the original feature space. Finally, $\tilde{\z}$ and $\bar{A}$ are fed into the decoder $p_{\zeta}(\mathbf{x}|\z,A)$ to compute the counterfactual instance.

\section{Root Cause Localization and Anomaly Mitigation (RootCLAM)}
\begin{figure*}[t]
    \centering
    \includegraphics[width=0.90\textwidth]{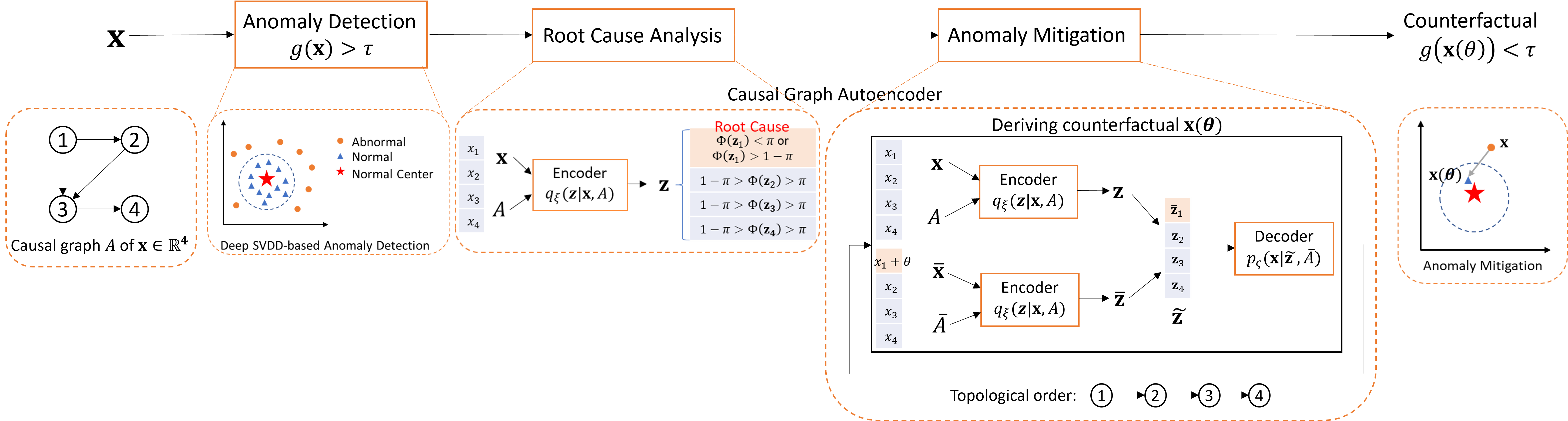}
    \caption{The pipeline to achieve root cause identification and anomaly mitigation.}
    \label{fig:framework}
\end{figure*}

In this section, we introduce RootCLAM, which is a two-phase framework that recommends anomaly mitigation actions to flip abnormal outcomes to normal ones. When an anomaly is detected, root cause localization is first to identify the abnormal features leading to the abnormal outcome. Then, anomaly mitigation is to further find actions on an anomaly to flip the prediction from a fixed anomaly detection model with the consideration of the root cause of the anomaly. Figure \ref{fig:framework} illustrates our framework for root cause analysis and anomaly mitigation.

\subsection{Problem Formulation}
We start with formulating the problem for root cause localization and anomaly mitigation. Consider an unlabeled dataset $\mathcal{X}=\{\mathbf{x}^{(n)}\}_{n=1}^N$ consisting of both normal and abnormal samples, where $\mathbf{x}=[x_1,...,x_i,...x_d] \in \mathbb{R}^d$ is a sample with $d$ features. We adopt a score-based anomaly detection model $g(\cdot):\mathcal{X} \rightarrow \mathbb{R}$, which labels abnormal samples if $g(\mathbf{x})>\tau$, where $\tau$ indicates the threshold. By applying $g(\cdot)$ on $\mathcal{X}$, we can obtain a set of detected abnormal samples $\hat{\mathcal{X}}^{-}$. Our goal is to find the root causes of the anomalies as well as the actions to fix them.

{\bf \noindent Root Cause}.
First, we need to define the root cause. Assume that the normal data are generated from a Structural Causal Model (SCM) given as follows:
\begin{equation*}
    \forall x_i\in \mathcal{X}, \quad x_i \sim P(x_i | \{ x_j, \forall j\in X_{\Pa{i}} \}, u_i).
\end{equation*}
We consider that any anomaly is caused by certain external interventions on some features in the SCM. Thus, the root causes of anomalies are defined as follows.
\begin{definition}
    %\label{def: root cause}
    Given any anomaly $\mathbf{x} \in \hat{\mathcal{X}}^{-}$, the root causes of $\mathbf{x}$ is a set of features $\mathcal{I}$ that receives external interventions.
\end{definition}
We do not assume the type of the SCM, but we do assume that the external intervention on a feature $x_i$ can be represented as an intervention on the exogenous variable $u_i$. It is straightforward to show that this assumption holds for some common types of SCM, such as the additive noise model where the structural function is a linear combination of $X_{\Pa{i}}$ and $u_{i}$. Based on this assumption, we treat the root cause as the feature where the intervention leads to a significant change in its distribution. 
\begin{definition}
    \label{def: root cause}
    (Root cause). Given an anomaly $\mathbf{x} \in \hat{\mathcal{X}}^{-}$, the root cause of $\mathbf{x}$ is a set of features $\mathcal{I}$ that receives an external intervention leading to a significant change in the marginal distributions of exogenous variables $P(u_\mathcal{I})$.
\end{definition}

It is worth noting that the features that are not the root cause may still exhibit abnormal behaviors. For example, suppose that a feature $x_i$ receives an external intervention, meaning that the probability distribution $P(u_i)$ is changed to a different distribution $P'(u_i)$. Meanwhile, the change in $x_i$ may propagate through the SCM, influencing another downstream feature $x_j$, where $x_j$ is a child of $x_i$ defined by SCM. As a result, the value of $x_j$ may also become abnormal due to the propagation from the external intervention on $x_i$ through the SCM, despite being a non-root cause.

{\bf \noindent Anomaly Mitigation}. Once the anomaly is detected, one can perform recourse actions to modify the values of certain features to change the abnormal sample to a normal one. As it is natural to modify root cause features only, we consider the problem of anomaly mitigation that asks to find a minimum perturbation on the root cause features $i \in \mathcal{I}$ of a sample to flip the label made by $g(\cdot)$. From the causal perspective, the recourse actions can be modeled as soft interventions. Specifically, define the anomaly mitigation action as a parameter vector $\theta=[\theta_1,...,\theta_i,...\theta_d]$ ($\theta_j=0$ if $j\notin \mathcal{I}$). For each root cause feature $x_i$, we formulate the action that changes $x_i$ to $x_i+\theta_i$ as a soft intervention. Then, the consequence of the action on a sample $\x$ is the counterfactual instance of $\x$ under the soft intervention. We denote this counterfactual instance as $\x(\theta)$ which depends on the value of $\theta$ as well as the underlying SCM.

With the above notations, the problem of anomaly mitigation becomes to find the parameter vector $\theta$ that minimizes the cost of the changes made by the mitigation actions, subject to making the counterfactual instance $\x(\theta)$ a normal sample for each original abnormal sample $\x$. 
It is formulated as that the anomaly detection model should have the anomaly score less than the threshold $\tau$ by taking counterfactual sample $\x(\theta)$ as input, i.e., $g(\x(\theta)) \leq \tau$.
By using the weighted L2 norm of the action values $\theta$ as the quantitative cost measure, given by $\|\mathbf{c}\cdot \theta\|_2$ where $\mathbf{c}$ is a cost vector for describing costs of revising all root cause features ($c_j=1$ if $j\notin \mathcal{I}$), the problem is finally formulated as
\begin{equation}
\label{eq:problem}
     \argmin_{\theta}  \|\mathbf{c} \cdot \theta\|_2 \quad
     \textrm{s.t. } \forall \x \in \hat{\mathcal{X}}^{-},  g(\x(\theta)) \leq \tau
\end{equation}

Solving the optimization problem in Eq.~\eqref{eq:problem} is not trivial. When an action is performed to change $x_i$ to $x_i+\theta_i$, the downstream features that are causally related will also be affected by this action. For example, changing an annual salary usually has an impact on the account balance. 
Thus, the counterfactual instance $\x(\theta)$ is not simply equal to $\x+\theta$. Ignoring causal relationships will lead to incorrect action recommendations, and counterfactual inference is needed to derive the accurate consequence of actions. 
Next, we address this challenge by leveraging the Variational Causal Graph Autoencoder (VACA), a state-of-the-art causal graph autoencoder.

\subsection{Root Cause Localization}
Based on the Definition \ref{def: root cause}, the idea of localizing the root cause features is to examine the exogenous variables of all features. If an exogenous variable $u_i$ does not follow the regular distribution $P(u_i)$ learned from the normal data, the exogenous variable should be the root cause of an anomaly that receives the external intervention. In this way, even if a feature is abnormal, as long as its exogenous variable follows a similar distribution as the normal data, we treat it as a non-root cause feature and attribute the abnormal behavior to be propagated from its parents.

To this end, we leverage VACA to learn the distribution of the exogenous variable. As mentioned earlier, VACA contains an encoder that maps the features to a hidden exogenous representation, i.e., $\z \sim q_{\xi}(\z| \x, A)$, as well as a decoder that maps the hidden exogenous representation back to the feature space, i.e., $\x \sim p_{\zeta}(\x|\z, A)$. The decoder and encoder are implemented as graph neural networks, and all computations follow the structural equation specified by the SCM. For each feature $x_i\in \x$, the purpose of $z_i\in \z$ is to capture the information of $x_i$ that cannot be explained by its parents. Thus, $z_i$ plays a similar role to $u_i$, which implies that we can examine the distribution of $\z$ to localize the root causes.

Specifically, after training the VACA on normal data, for each sample $\mathbf{x} \in \hat{\mathcal{X}}^{-}$, we first derive the hidden variable $\mathbf{z}$ based on the encoder of VACA and further calculate the cumulative probability $\Phi(z_i)$ for each exogenous variable based on the distribution fitted from normal data. To identify the root cause features with significant changes in exogenous variables, we set a threshold $\pi$ for the percentage of the values (in our experiments we use $\pi=0.125$). If $\Phi(z_i)$ is smaller than $\pi$ or larger than $1-\pi$, we consider the feature $x_i$ as a potential root cause. As there can be multiple root cause features in a particular sample, we examine the exogenous variables of all features and get a set of root cause features $\mathcal{I}$.   

\subsection{Causal Graph Autoencoder-based Anomaly Mitigation}
For each sample in $\hat{\mathcal{X}}^{-}$, after getting the root causes, we further want to flip the abnormal outcome with minimum actions on root cause features $\mathcal{I}$. 
The challenge in solving Eq.~\eqref{eq:problem} is how to compute counterfactual instance $\x(\theta)$ and solve $\theta$ as a continuous optimization problem. We propose to perform the Abduction-Action-Prediction process to conduct the counterfactual inference based on the VACA. Since we perform actions on all features, we consider an iterative Abduction-Action-Prediction process as follows:
\begin{equation}\label{eq:xi}
\begin{split}
    & x_1(\theta) = \underbrace{x_1 + \theta_1}_{\textrm{Action}}, \\
    & \textrm{for } i=2\cdots d, \quad \tilde{x}_i \sim \underbrace{ \int P(x_i | \{ x_j(\theta), \forall j\in \Pa{i} \} , u_i ) \underbrace{P(u_i|\x)}_{\textrm{Abduction}}  \mathrm{d}\tilde{u}_i }_{\textrm{Prediction}}, \\
    & \quad\quad\quad\quad\quad\quad\quad x_{i}(\theta) = \underbrace{\tilde{x}_i + \theta_i}_{\textrm{Action}},
\end{split}
\end{equation}
where the features are sorted in topological order. More specifically, to compute $\x(\theta)$, we: (1) infer the updated probability $P(u_i|\x)$ (Abduction); (2) perform the action on each feature $x_i$ (Action); and (3) infer the counterfactual values of the downstream features. Steps (2) and (3) are repeated until all features are modified.

There are two challenges in directly applying the VACA to our context. First, the VACA is designed to perform hard intervention where the connections from the parents to the intervened node are cut off. However, in our context, we conduct interventions on all actionable features. By using hard intervention, the parent-child relations of multiple features would be cut-off and cannot pass to downstream nodes, which totally changes the underlying SCM making the generated counterfactual instances infidelity. Therefore, we perform soft interventions on all features where the parent-child relations are preserved, which cannot be achieved by directly using the VACA to perform hard interventions on all features. Second, the hidden exogenous representation $\z$ produced by the encoder may not be in the same space as the features, but we want to compute the recourse on the original feature space. These two challenges mean that the action values cannot be directly added on $\z$ when we adopt the VACA as the causal graph autoencoder.

We address the above challenges by proposing an iterative algorithm, where each iteration performs a hard intervention on one feature following a topological order. The idea is to pass the influence of each hard intervention to the downstream nodes before performing the hard intervention on the next node in the topological order, in order to simulate how the soft intervention works.
Specifically, at the $i$th iteration, to take the generated action on feature $X_i$, we perform a hard intervention on $X_i$ as $do(X_i=x_i+\theta_i)$ to obtain the intervened instance $\bar{\x}$. Then, we use the VACA to compute the interventional influence on all descendants of $X_i$ similarly to the above discussion. In this process, $\bar{\x}$ is first transformed to the hidden representation $\bar{\z}$ by the encoder. Meanwhile, the sample $\x$ before the intervention is also transformed to the hidden representation $\z$ by the encoder. Then, $\bar{z}_i$ in $\bar{\z}$ replaces $z_i$ in $\z$ to perform the intervention in the hidden space that is equivalent to performing the intervention in the original feature space. Finally, the interventional influences of this action are transmitted to all descendants of $X_i$ by the decoder which produces the counterfactual instance of the sample under the intervention. It is worth noting that, at the beginning of the $i$th iteration, the value of $x_i$ has already been updated by taking into account the interventional influences of actions taken on ancestors of $X_i$. As a result, after we perform the hard intervention on all features, we obtain the counterfactual instance under the recourse. 

\begin{algorithm}
\caption{Training Procedure of RootCLAM for Mitigation Action Prediction}\label{alg:1}
    \ForEach{$\x \in \hat{\mathcal{X}}^{-}$}{
        Compute root cause features $\mathcal{I}$ for $\x$ \\
        $\tilde{\x} \leftarrow \x$ \\
    	\ForEach{$i \in \mathcal{I}$}{
                    Compute $\theta_i = h_{\phi_i}(\x)$   \\ 
        	    Draw $\tilde{\z} \sim q_{\xi}(\z| \tilde{\x}, A)$ \hfill\tcp{Abduction} 
        	    Compute $\bar{x}_i(\theta) = \tilde{x}_i + \theta_i$ \hfill\tcp{Action}
        	    Replace $\tilde{x}_i$ in $\x$ with $\bar{x}_i(\theta)$ and get $\bar{\x}$ \\
        	    Draw $\bar{\z} \sim q_{\xi}(\z|\bar{\x}, \bar{A})$ \\
        	    Replace $\tilde{z}_i$ in $\tilde{\z}$ with $\bar{z}_i$ in $\bar{\z}$ and get $\z(\theta)$ \\
        	    Draw $\x(\theta) \sim p_{\zeta}(\x|\z(\theta), \bar{A})$ \hfill\tcp{Prediction} 
        	    $\tilde{\x} \leftarrow \x(\theta)$  \\      
    	} 
    	Compute $\mathcal{L}(\phi)$ according to Eq.~\eqref{eq:obj} \\
    	Compute $\frac{\partial \mathcal{L}(\phi)}{\partial \phi}$ \\ 
    	Update $\phi = \phi - \eta \frac{\partial \mathcal{L}(\phi)}{\partial \phi}$  
	} 
    \Return $h_{\phi}$ 
\end{algorithm}

Finally, for the sake of generalization, instead of computing $\theta$ for each instance separately, we define a function $\theta = h_{\phi}(\x)$ for generating the action given $\x$. By integrating the score-based anomaly detection model and VACA for computing the counterfactual instance into Eq.~\eqref{eq:problem} and adding the constraint to the objective as regularization, we obtain the final objective function as follows:
\begin{equation}
\label{eq:obj}
    \mathcal{L}(\phi) = \!\!\!\!\! \sum_{\x^{(n)} \in \hat{\mathcal{X}}^{-}} \!\!\!\! \max\left\{g\left(\x^{(n)}(\theta^{(n)})\right)-\alpha \tau, 0\right\} 
     + \lambda  \|\mathbf{c}\cdot \theta^{(n)} \|_2,
\end{equation}
where $\theta^{(n)}=h_{\phi}(\x^{(n)})$ indicates the action values for the sample $\x^{(n)}$; $\lambda$ is a hyperparameter balancing the actions on the anomalies and the flipping of abnormal outcomes; $\alpha$ is another hyperparameter controlling how close the anomaly score of counterfactual sample should be to the threshold $\tau$. Note that the only trainable parameters in this objective function are the parameters $\phi$ of $h_{\phi}(\x)$ for generating the action values. Eq.~\eqref{eq:obj} can be minimized using off-the-shelf gradient-based optimization algorithms. The training procedure is shown in Algorithm \ref{alg:1}.

{\bf \noindent Practical Considerations.} RootCLAM assumes the availability of a causal graph about the data. In practice, the causal graphs may not be available. In this case, we can leverage the causal discovery algorithms to identify the causal relations of observational data \cite{glymour2019review}.

\section{Experiments}

\subsection{Experimental Setup}

{\bf \noindent Datasets.} We conduct experiments on two semi-synthetic datasets and one real-world dataset. For the real-world dataset, as we do not have the ground-truth SCM, we only use it for a case study.

$\bullet$ \textbf{Loan} \cite{karimi2020algorithmic} is a \textit{semi-synthetic dataset} about a loan approval scenario derived from the German Credit dataset \cite{Dua:2019}, which consists of 7 endogenous features including loan amount (L), loan duration (D), income (I), savings (S), education level (E), age (A), and gender (G). The label Y indicates the probability of loan approval.
We treat the samples with high approval probabilities as normal and the samples with low approval probabilities as anomalous. The structural equations for data generation are be found in \cite{karimi2020algorithmic}. Due to the space limit, we do not include the equations in this paper. 
% \begin{equation*} 
%     \begin{split}
%     \label{eq:loan_x}
%         f_G: G &= U_G \\ 
%         f_A: A &= -35+U_A \\
%         f_E: E &=-0.5+(1+e^{+1-0.5G-(1+e^{-0.1A})^{-1}}-U_E)^{-1} \\
%         f_L: L &=1+0.01(A-5)(5-A)+G+U_L \\
%         f_D: D &=-1+0.1A+2G+L+U_D \\
%         f_I: I &=-4+0.1(A+35)+2G+GE+U_I \\
%         f_S: S &=-4+1.5\mathbbm{1}_{I>0}I+U_S
%     \end{split}
% \end{equation*}
% where $U_G \sim Bernoulli(0.5)$, $U_A \sim Gamma(10,3.5)$, $U_E \sim \mathcal{N}(0,0.25)$, $U_L \sim \mathcal{N}(0,4)$, $U_D \sim \mathcal{N}(0,9)$, $U_S \sim \mathcal{N}(0,25)$, $U_I \sim \mathcal{N}(0,4)$.
% The labels $Y$ are sampled based on 
% \begin{equation*}
% \label{eq:loan_y}
%     Y = (1+e^{-0.3(-L-D+I+S+IS)})^{-1}.
% \end{equation*}

$\bullet$ \textbf{Adult} \cite{sanchez2021vaca} is another \textit{semi-synthetic dataset} about the annual income of a person derived from the real-world Adult dataset \cite{Dua:2019}, which consists of 10 endogenous features of a person including age (A), education level (E), hours worked per week (H), race (R), native country (N), sex (S), work status (W), marital status (M), occupation sector (O), and relationship status (L). We use the SCM designed in the paper \cite{sanchez2021vaca}. We follow the common settings of the adult dataset to treat samples with income less than \$50k as normal and samples with income more than \$50k as abnormal. We use the structural equations for data generation defined in \cite{sanchez2021vaca}. 

{\bf \noindent Anomaly Injection.}
To quantify the performance of RootCLAM for root cause localization, we generate abnormal samples by revising exogenous variables of some features. Especially, to generate anomalies, we first randomly select one to four features and then change the distribution of the corresponding exogenous variables. For example, on the Loan dataset, we change the exogenous variable $U_S$ of savings (S) from $\mathcal{N}(0,25)$ to $\mathcal{N}(-25,25)$. In this way, we have the ground truth of the root causes for each abnormal sample.

$\bullet$ \textbf{Donors}~\footnote{https://www.kaggle.com/c/kdd-cup-2014-predicting-excitement-at-donors-choose} is a \textit{real-world dataset} that aims to predict whether a project on DonorsChoose.org is exciting to the business. The dataset consists of 10 endogenous features of a project, including ``at least one teacher-referred donor'', ``fully funded'', ``at least one green donation'', ``great chat'', ``three or more non teacher-referred donors'', ``one non teacher-referred donor giving 100 plus'', ``donation from thoughtful donor'', ``great messages proportion'', ``teacher-referred count'', ``non teacher-referred count''. A project must meet all of the following five criteria to be exciting: 1) was fully funded; 2) had at least one teacher-referred donor; 3) has a higher than average percentage of donors leaving an original message; 4) has at least one ``green'' donation; 5) has one or more of: 5.1) donations from three or more non teacher-referred donors, 5.2) one non teacher-referred donor gave more than \$100, 5.3) the project received a donation from a ``thoughtful donor''.

\begin{figure}[h!]
    \centering
    \includegraphics[width=0.40\textwidth]{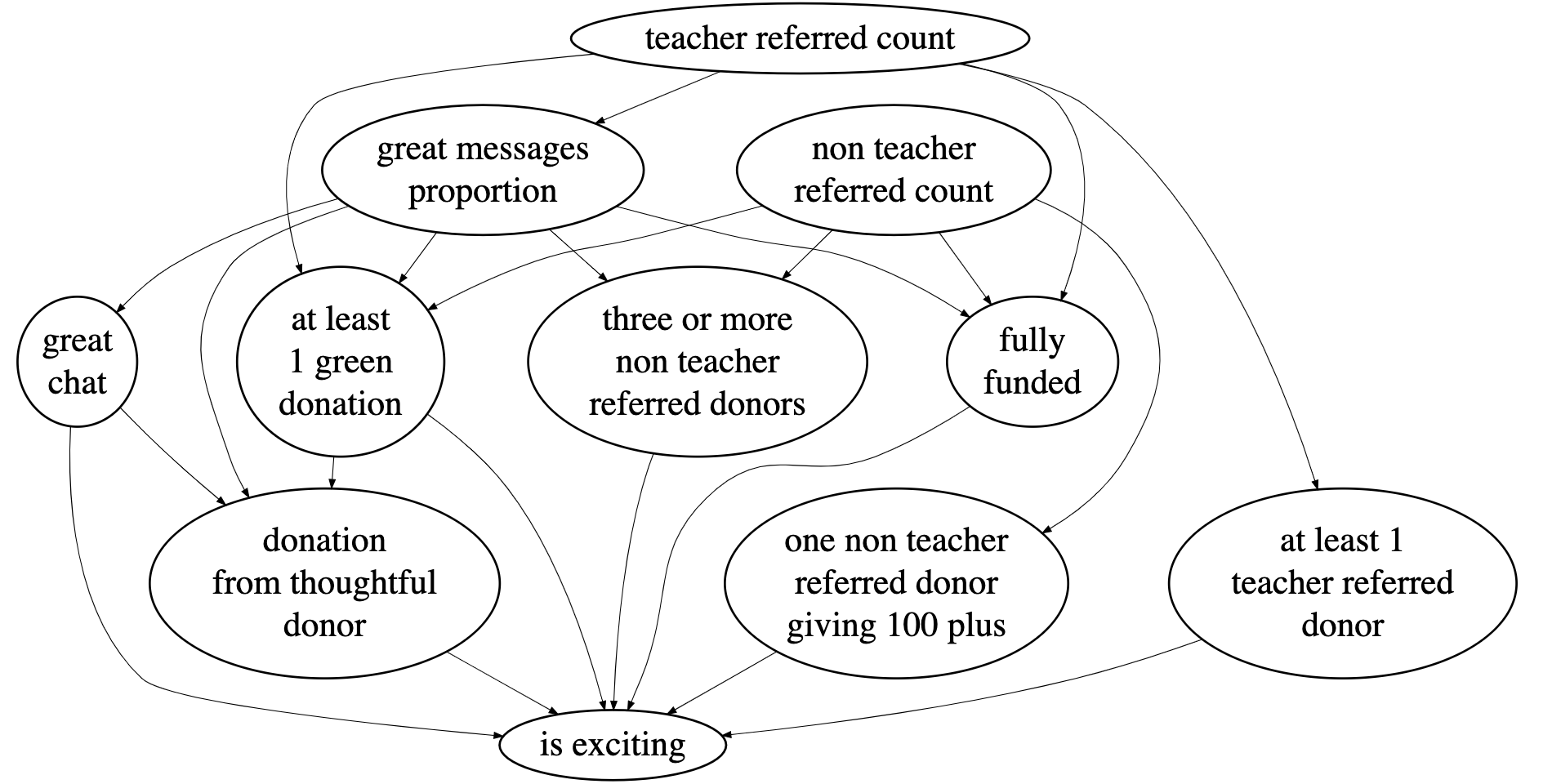}
    \caption{Learned causal graph on Donors.}
    \label{fig:graph_donors}
\end{figure}

We consider exciting projects as normal and non-exciting projects as abnormal, while anomaly mitigation is to provide guidance to make the project exciting. As a real-world dataset, we do not have the ground-truth SCM, so we only use it for a case study. The causal graph used in RootCLAM is approximated by the PC algorithm \cite{kalisch2007estimating} with some minor edits to incorporate the domain knowledge. Figure \ref{fig:graph_donors} shows the causal graph on Donors.

% \begin{figure}[h!]
%     \centering
%     \includegraphics[width=0.48\textwidth]{figures/donors_graph.png}
%     \caption{Learned causal graph on Donors.}
%     \label{fig:graph_donors}
% \end{figure}

% \begin{figure}[h!]
%     \centering
%     \includegraphics[width=0.48\textwidth]{figures/graphviz.png}
%     \caption{Learned causal graph on Donors.}
%     \label{fig:graph_donors}
% \end{figure}

Table \ref{tb:datasets} shows the statistics of three datasets. To simulate the anomaly detection scenario, we set the ratio of abnormal samples to normal samples as 1:10 in the unlabeled dataset for testing.

\begin{table}
\small
\caption{Statistics of three datasets.}
% The numbers in the parentheses indicate \# of categorical feature and \# of continuous feature.}
\label{tb:datasets}
\centering
\begin{tabular}{|c|c|c|cc|}
\hline
\multirow{2}{*}{Dataset} & \multirow{2}{*}{\# of Features} & \multirow{2}{*}{Normal Dataset} & \multicolumn{2}{c|}{Unlabeled Dataset}  \\ \cline{4-5} 
                         &                           &                                 & \multicolumn{1}{c|}{Normal} & Anomalous \\ \hline
Loan                     & 7                        & 10,000                          & \multicolumn{1}{c|}{10,000} & 1,000      \\ \hline
Adult                    & 10                        & 10,000                          & \multicolumn{1}{c|}{10,000} & 1,000      \\ \hline
Donors                   & 10                        & 10,000                          & \multicolumn{1}{c|}{26,710} & 2,671      \\ \hline
\end{tabular}
\end{table}
%Loan (1/6); Adult (7/3); Donors (7/3)

{\bf \noindent Anomaly Detection Models.} We adopt Deep Support Vector Data Description (Deep SVDD) \cite{ruff2018deep} and autoencoder-based model (AE) \cite{ruff2021unifying} as anomaly detection models $g(\cdot)$. 

$\bullet$ \textbf{Deep SVDD} derives the anomaly scores of the test sample based on its distance to the center $\boldsymbol\mu$ of a hypersphere constructed by normal samples, i.e., $g(\x) = \|r(\x)-\boldsymbol\mu\|_2$, where $r(\x)$ indicates the hidden representation of a sample $\x$ derived from $r(\cdot)$. Then, the objective function (Eq.~\eqref{eq:obj}) for the recourse recommendation can be rewritten as:
\begin{equation*}
    \mathcal{L}_{S}(\phi)\!\! = \!\!\!\!\sum_{\x^{(n)} \in \hat{\mathcal{X}}^{-}}\!\!\! \max\{\|\x^{(n)}(\theta^{(n)}) -\boldsymbol\mu\|_2-\alpha \tau, 0\} + \lambda  \|\mathbf{c}\cdot \theta^{(n)} \|_2.
\end{equation*}

$\bullet$ \textbf{AE}-based anomaly detection model derives the anomaly scores of samples based on the reconstruction errors of an autoencoder that is trained by normal samples, i.e., $g(\x) = \|\x-\hat\x\|_2$, where $\hat\x$ indicates the reconstructed sample from autoencoder. Then, to provide recourse for the AE-based anomaly detection model, the objective function (Eq.~\eqref{eq:obj}) can be rewritten as:
\begin{equation*}
% \begin{split}
    \mathcal{L}_{AE}(\phi) = \!\!\!\!
      \sum_{\x^{(n)} \in \hat{\mathcal{X}}^{-}}\!\!\! \max\{\|\x^{(n)}(\theta^{(n)}) \!-\!\widehat{\x^{(n)}(\theta^{(n)})}\|_2\!-\!\alpha \tau, 0\}  + \lambda  \|\mathbf{c}\cdot \theta^{(n)} \|_2.
% \end{split}
\end{equation*}

In our experiments, we first train Deep SVDD and AE on the normal dataset, respectively, and then apply the models on the unlabeled dataset $\mathcal{X}$ and get the corresponding $\hat{\mathcal{X}}^{-}$ from each model.

{\bf \noindent Baseline for Root Cause Localization.} We compare RootCLAM with CausalRCA \cite{janzing2019causal}, a state-of-the-art approach for root cause analysis. We use the implementation in the DoWhy package \cite{dowhypaper}.

{\bf \noindent Baselines for Anomaly Mitigation.} To our best knowledge, there is no causal anomaly mitigation approach. We compare RootCLAM with two baselines, C-CHVAE and NaiveAM. 

$\bullet$ \textbf{C-CHVAE} \cite{pawelczyk2020learning} can find feasible counterfactual flipping the output of classifiers, but does not consider the underlying causal relationships when generating counterfactuals. We adapt C-CHVAE by replacing classifiers with anomaly detection models. 

$\bullet$ \textbf{NaiveAM} directly predicts the action values on all feasible features without considering the underlying causal structure. Specifically, given a set of abnormal sample $\hat{\mathcal{X}}^{-}$, we still train a neural network $\hat{h}_\phi(\cdot)$ to predict the action value, $\hat\theta = \hat{h}_\phi(\x)$, where $\x \in \hat{\mathcal{X}}^{-}$. However, instead of generating the counterfactual samples guided by SCM, NaiveAM generates the revised samples by simply adding the action value on the original sample, i.e., 
\begin{equation}
    \label{eq:baseline}
    \hat{\x}(\theta) = \x+\hat\theta.
\end{equation} 
NaiveAM is also trained on the objective function in Eq.~\eqref{eq:obj} by replacing $\theta$ and $\x(\theta)$ with $\hat\theta$ and $\hat{\x}(\theta)$, respectively. After training, in order to evaluate whether the predicted actions can really flip the labels in the counterfactual world, on Adult and Loan datasets, we also generate the counterfactual samples based on the structural equations given $\hat\theta$, denoted as $\hat{\x}(\theta)$ (SCM). 

{\bf \noindent Implementation Details.} For a fair comparison, the hyperparameters of neural networks for action prediction in NaiveAM and RootCLAM are the same. We set the hyperparameters for VACA by following \cite{sanchez2021vaca}. By default, the threshold for anomaly detection is set to 0.995 quantiles of the training samples' distances to the center (Deep SVDD) or the reconstruction errors (AE). For the intervention value prediction, we utilize a feed-forward network with structure m-2048-2048-n, where m is the input dimension and n is the number of actionable features. The costs $\mathbf{c}$ in Eq.~\eqref{eq:obj} are user-specified functions for each root cause feature to represent preferences or feasibility of features changing. The cost functions can be changed according to the requirements or prior knowledge. To be fair, we use the standard deviation of each root cause feature as the cost for NaiveAM and RootCLAM. Our code is available online \footnote{https://github.com/hanxiao0607/RootCLAM}.

\subsection{Experimental Results}

{\bf \noindent The performance of anomaly detection.} We evaluate the performance of anomaly detection in terms of the F1 score, the area under the receiver operating characteristic (AUROC), and the area under the precision-recall curve (AUPRC). Table \ref{tb:ad} shows the anomaly detection evaluation results. In short, both AE and Deep SVDD can achieve good performance for anomaly detection, meaning that the predicted abnormal samples $\tilde{\mathcal{X}}^-$ have high accuracy. It lays a solid foundation for action prediction.

After getting the abnormal set $\tilde{\mathcal{X}}^-$ of each dataset, we then train and test the root cause localization and anomaly mitigation with the train/test split ratio of 80/20.

\begin{table}[ht]
\centering
\caption{Anomaly detection on the unlabeled datasets.}
\label{tb:ad}
\resizebox{0.47\textwidth}{!}
{
\begin{tabular}{|c|ccc|ccc|}
\hline
\multirow{2}{*}{Dataset} & \multicolumn{3}{c|}{AE}                                            & \multicolumn{3}{c|}{Deep SVDD}                                     \\ \cline{2-7} 
                         & \multicolumn{1}{c|}{F1}     & \multicolumn{1}{c|}{AUROC}  & AUPRC  & \multicolumn{1}{c|}{F1}     & \multicolumn{1}{c|}{AUROC}  & AUPRC  \\ \hline
Loan                     & \multicolumn{1}{c|}{0.923} & \multicolumn{1}{c|}{0.998} & 0.982 & \multicolumn{1}{c|}{0.888} & \multicolumn{1}{c|}{0.993} & 0.944 \\ \hline
Adult                    & \multicolumn{1}{c|}{0.893} & \multicolumn{1}{c|}{0.984} & 0.899 & \multicolumn{1}{c|}{0.837} & \multicolumn{1}{c|}{0.923} & 0.823 \\ \hline
Donors                   & \multicolumn{1}{c|}{0.967} & \multicolumn{1}{c|}{0.998} & 0.979 & \multicolumn{1}{c|}{0.988} & \multicolumn{1}{c|}{0.999} & 0.998 \\ \hline
\end{tabular}
}
\end{table}

\begin{table}[h]
\caption{Root cause localization on the unlabeled datasets.}
\label{tab:rc}
\resizebox{0.47\textwidth}{!}{
\begin{tabular}{|c|c|cccc|cccc|}
\hline
\multirow{2}{*}{}      & \multirow{2}{*}{} & \multicolumn{4}{c|}{AE}                                                                      & \multicolumn{4}{c|}{Deep SVDD}                                                               \\ \cline{3-10} 
                       &                   & \multicolumn{1}{c|}{Accu.} & \multicolumn{1}{c|}{Pre.}  & \multicolumn{1}{c|}{Rec.}  & F1    & \multicolumn{1}{c|}{Accu.} & \multicolumn{1}{c|}{Pre.}  & \multicolumn{1}{c|}{Rec.}  & F1    \\ \hline
\multirow{2}{*}{Loan}  & CausalRCA         & \multicolumn{1}{c|}{0.707} & \multicolumn{1}{c|}{0.522} & \multicolumn{1}{c|}{0.680} & 0.591 & \multicolumn{1}{c|}{0.704} & \multicolumn{1}{c|}{0.508} & \multicolumn{1}{c|}{0.561} & 0.533 \\ \cline{2-10} 
                       & RootCLAM          & \multicolumn{1}{c|}{0.728} & \multicolumn{1}{c|}{0.545} & \multicolumn{1}{c|}{0.765} & 0.636 & \multicolumn{1}{c|}{0.727} & \multicolumn{1}{c|}{0.523} & \multicolumn{1}{c|}{0.776} & 0.631 \\ \hline \hline
\multirow{2}{*}{Adult} & CausalRCA         & \multicolumn{1}{c|}{0.853} & \multicolumn{1}{c|}{0.554} & \multicolumn{1}{c|}{0.615} & 0.583 & \multicolumn{1}{c|}{0.850} & \multicolumn{1}{c|}{0.546} & \multicolumn{1}{c|}{0.593} & 0.569 \\ \cline{2-10} 
                       & RootCLAM          & \multicolumn{1}{c|}{0.866} & \multicolumn{1}{c|}{0.567} & \multicolumn{1}{c|}{0.849} & 0.680 & \multicolumn{1}{c|}{0.855} & \multicolumn{1}{c|}{0.544} & \multicolumn{1}{c|}{0.794} & 0.646 \\ \hline
\end{tabular}
}
\end{table}

\begin{table*}[ht]
\small
\caption{The performance of anomaly mitigation in terms of the flipping ratio and norm of action values.}
\label{tab:flipr}
\centering
% \resizebox{0.98\textwidth}{!}{
\renewcommand{\arraystretch}{1.15}
\begin{tabular}{|c|c|c|ccc|ccc|}
\hline
\multirow{2}{*}{}          & \multirow{2}{*}{Metric}         & \multirow{2}{*}{}              & \multicolumn{3}{c|}{Loan}                                                                                  & \multicolumn{3}{c|}{Adult}                                                                                    \\ \cline{4-9} 
                           &                                 &                                & \multicolumn{1}{c|}{C-CHVAE}            & \multicolumn{1}{c|}{NaiveAM}            & RootCLAM                  & \multicolumn{1}{c|}{C-CHVAE}              & \multicolumn{1}{c|}{NaiveAM}            & RootCLAM                   \\ \hline \hline
\multirow{3}{*}{AE}        & \multirow{2}{*}{Flipping Ratio} & $\mathbf{\hat{Y}}$             
                           & \multicolumn{1}{c|}{$1.000$}  & \multicolumn{1}{c|}{$1.000$}  & $0.891$      
                           & \multicolumn{1}{c|}{$0.114$}    & \multicolumn{1}{c|}{$0.885$}  & $0.960$       \\ \cline{3-9} 
                           &                                 & $\mathbf{Y}$                   
                           & \multicolumn{1}{c|}{$0.499$}  & \multicolumn{1}{c|}{$0.337$}  & $0.839$  
                           & \multicolumn{1}{c|}{$0.065$}    & \multicolumn{1}{c|}{$0.598$}  & $1.000$  \\ \cline{2-9} 
                           & Action Value                    & $\|\mathbf{c}\cdot \theta\|_2$ 
                           & \multicolumn{1}{c|}{$22.383$} & \multicolumn{1}{c|}{$6.382$} & $5.185$ 
                           & \multicolumn{1}{c|}{$115.862$} & \multicolumn{1}{c|}{$34.389$} & $14.504$  \\ \hline \hline
\multirow{3}{*}{Deep SVDD} & \multirow{2}{*}{Flipping Ratio} & $\mathbf{\hat{Y}}$             
                           & \multicolumn{1}{c|}{$1.000$}  & \multicolumn{1}{c|}{$1.000$}  & $0.988$      
                           & \multicolumn{1}{c|}{$0.671$}    & \multicolumn{1}{c|}{$1.000$}  & $1.000$       \\ \cline{3-9} 
                           &                                 & $\mathbf{Y}$                   
                           & \multicolumn{1}{c|}{$0.496$}  & \multicolumn{1}{c|}{$0.847$}  & $0.963$ 
                           & \multicolumn{1}{c|}{$0.586$}    & \multicolumn{1}{c|}{$0.595$}  & $1.000$   \\ \cline{2-9} 
                           & Action Value                    & $\|\mathbf{c}\cdot \theta\|_2$ 
                           & \multicolumn{1}{c|}{$17.832$} & \multicolumn{1}{c|}{$13.474$} & $5.862$ 
                           & \multicolumn{1}{c|}{$63.124$}   & \multicolumn{1}{c|}{$69.169$} & $29.274$ \\ \hline

\end{tabular}
% }
% \footnotesize{Significantly outperforms NaiveAM at the: * 0.05 or ** 0.01 level, paired t-test.}
\end{table*}

{\bf \noindent The performance of RootCLAM on root cause localization.}
After detecting the anomalies, the next step is to identify the root causes. We further evaluate the performance of RootCLAM on root cause localization in terms of accuracy, precision, recall, and F1. As shown in Table \ref{tab:rc}, RootCLAM outperforms CausalRCA in terms of accuracy and F1 score on both datasets. Especially, RootCLAM achieves much higher recall compared with CausalRCA, which means RootCLAM can identify more root cause features.

{\bf \noindent The performance of RootCLAM on counterfactual sample generation.}
Generating high-fidelity counterfactual samples is a fundamental requirement for predicting high-quality actions to flip the labels. We evaluate the quality of estimated counterfactual samples in terms of the mean squared error (MSE) as well as the standard deviation of the squared error (SSE) between the true and the estimated counterfactual samples on the Loan and Adult datasets that have the ground truth structural equations for data generation. On Loan, the MSE and SSE are 3.976 and 2.266, respectively, while on Adult, the MSE and SSE are 3.334 and 0.900, respectively. It means RootCLAM can get good counterfactual samples. 

{\bf \noindent The performance of anomaly mitigation in terms of flipping ratio.}
We evaluate the performance of anomaly mitigation by examining the flipping ratio that anomalies are transferred to normal through the interventions predicted by $h_\phi(\cdot)$. The flipping ratio is calculated as the fraction of the number of flipped samples over all detected anomalies. Because we would like to check whether the predicted actions can really flip the labels in the counterfactual world, given the predicted action values from RootCLAM and baselines, we also use the ground-truth structural equations to generate the counterfactual samples. We calculate the flipping ratio by considering two scenarios: 1) whether the anomaly detection model would detect the counterfactual samples as normal, denoted as $\hat{\mathbf{Y}}$; 2) whether the ground truth Y is flipping from abnormal to normal based on the ground-truth structural equations, denoted as $\mathbf{Y}$. 

As shown in Table \ref{tab:flipr}, on Loan and Adult datasets, both RootCLAM and NaiveAM can successfully flip almost all abnormal samples detected. However, C-CHVAE cannot get good performance on the Adult dataset.
For the flipping ratio on the ground truth label $\mathbf{Y}$, RootCLAM can successfully flip most of the abnormal samples on both datasets. It means the actions predicted by RootCLAM can reverse the majority of abnormal samples to normal in the counterfactual world. However, NaiveAM and C-CHVAE cannot get good performance on flipping the ground truth label $\mathbf{Y}$. This is because both NaiveAM and C-CHVAE do not consider the underlying causal structure in the data, showing that simply revising the root cause features is not sufficient to flip the ground-truth labels. 

{\bf \noindent The performance of anomaly mitigation in terms of the norm of action values.}
One requirement for anomaly mitigation is to conduct minimal interventions on the original samples. We further calculate the norm of action values, i.e., $\|\mathbf{c}\cdot \theta\|_2$, on the samples with successfully flipping labels. As shown in the last row of Table \ref{tab:flipr}, RootCLAM makes much smaller changes on the original samples  and still has higher flipping ratios on the ground truth label $\mathbf{Y}$.
% which shows the advantages of considering the causal relationships between features for anomaly mitigation.
% $\|x_{cf} - x\|_2$

\begin{figure}[t!]
\centering
    \begin{subfigure}{.23\textwidth}
      \centering
      \includegraphics[width=0.98\textwidth]{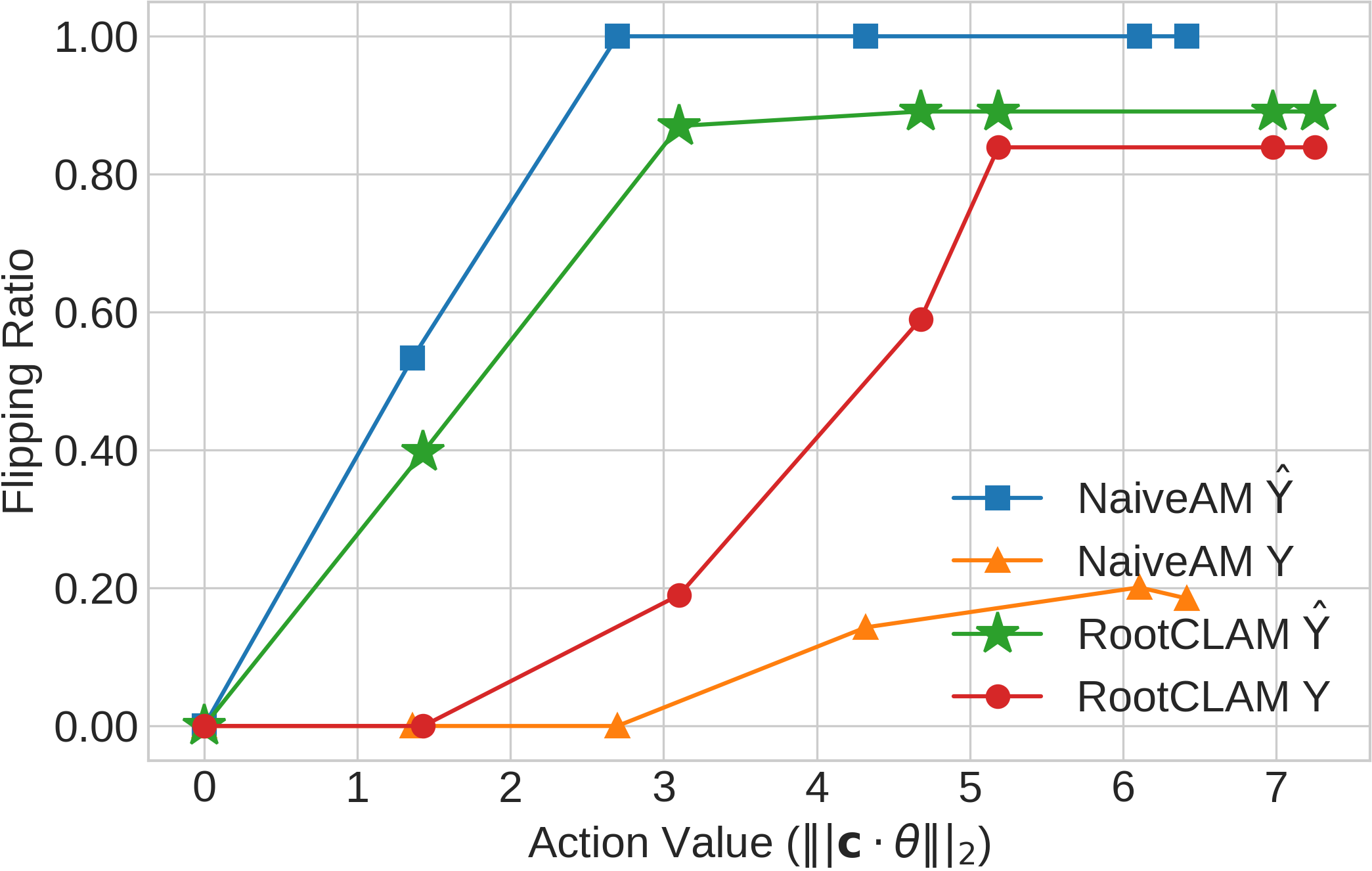}
      \caption{AE-Loan}
      \label{fig:loan_scm_ae}
    \end{subfigure}%
    \hfill
    \begin{subfigure}{.23\textwidth}
      \centering
      \includegraphics[width=0.98\textwidth]{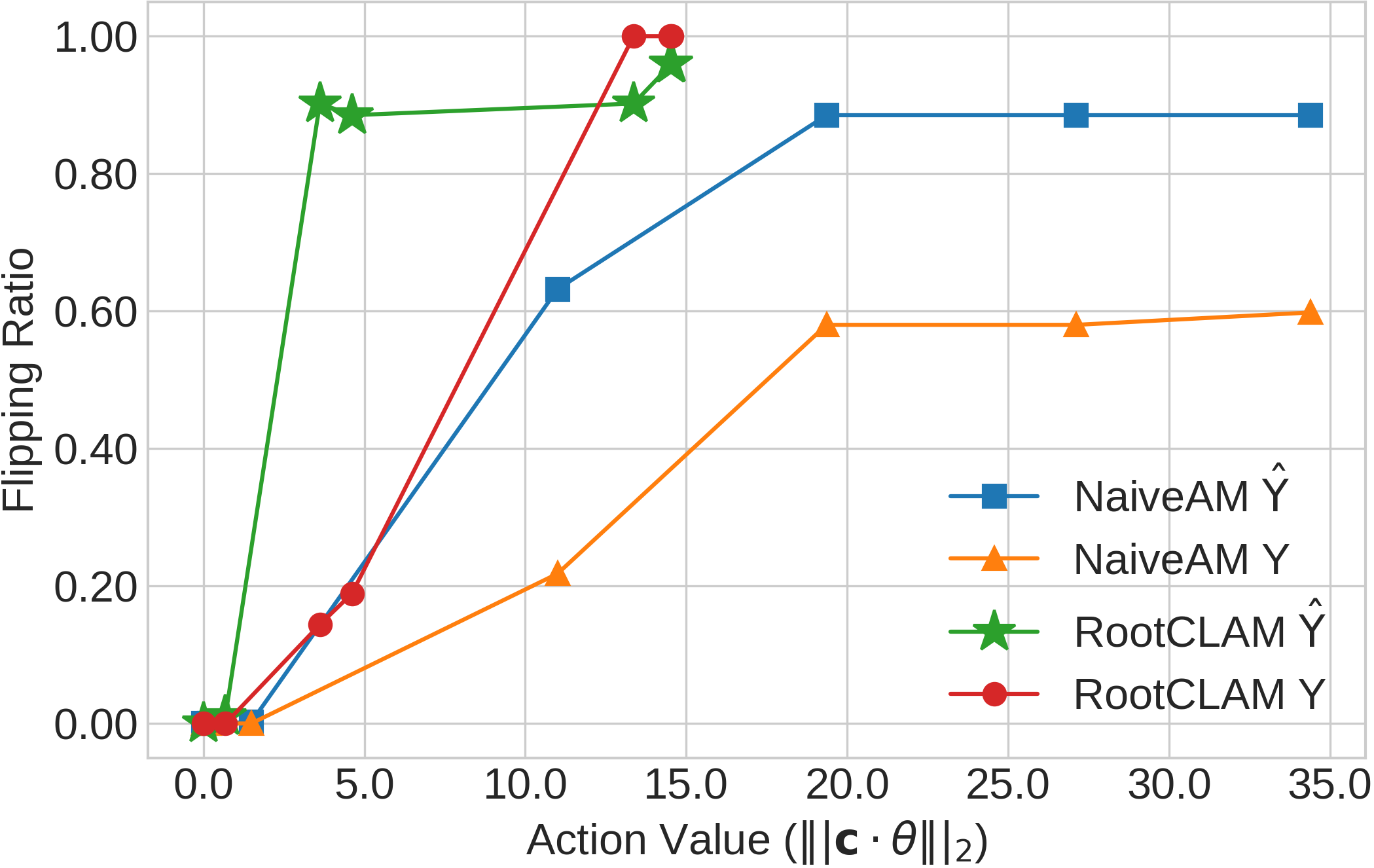}
      \caption{AE-Adult}
      \label{fig:adult_scm_ae}
    \end{subfigure}%
    \hfill
    \begin{subfigure}{.23\textwidth}
      \centering
      \includegraphics[width=0.98\textwidth]{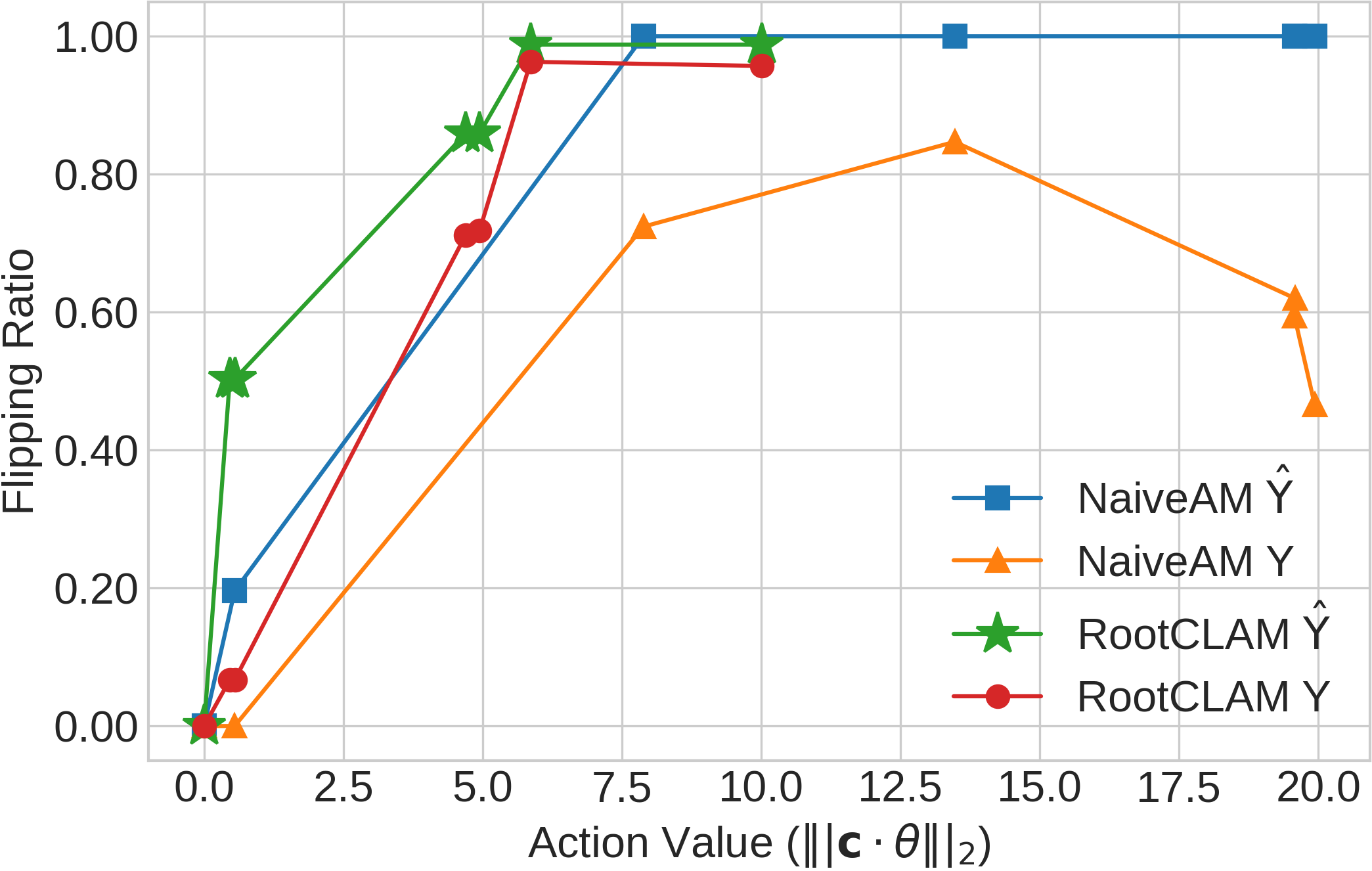}
      \caption{DeepSVDD-Loan}
      \label{fig:loan_scm}
    \end{subfigure}%
    \hfill
    \begin{subfigure}{.23\textwidth}
      \centering
      \includegraphics[width=0.98\textwidth]{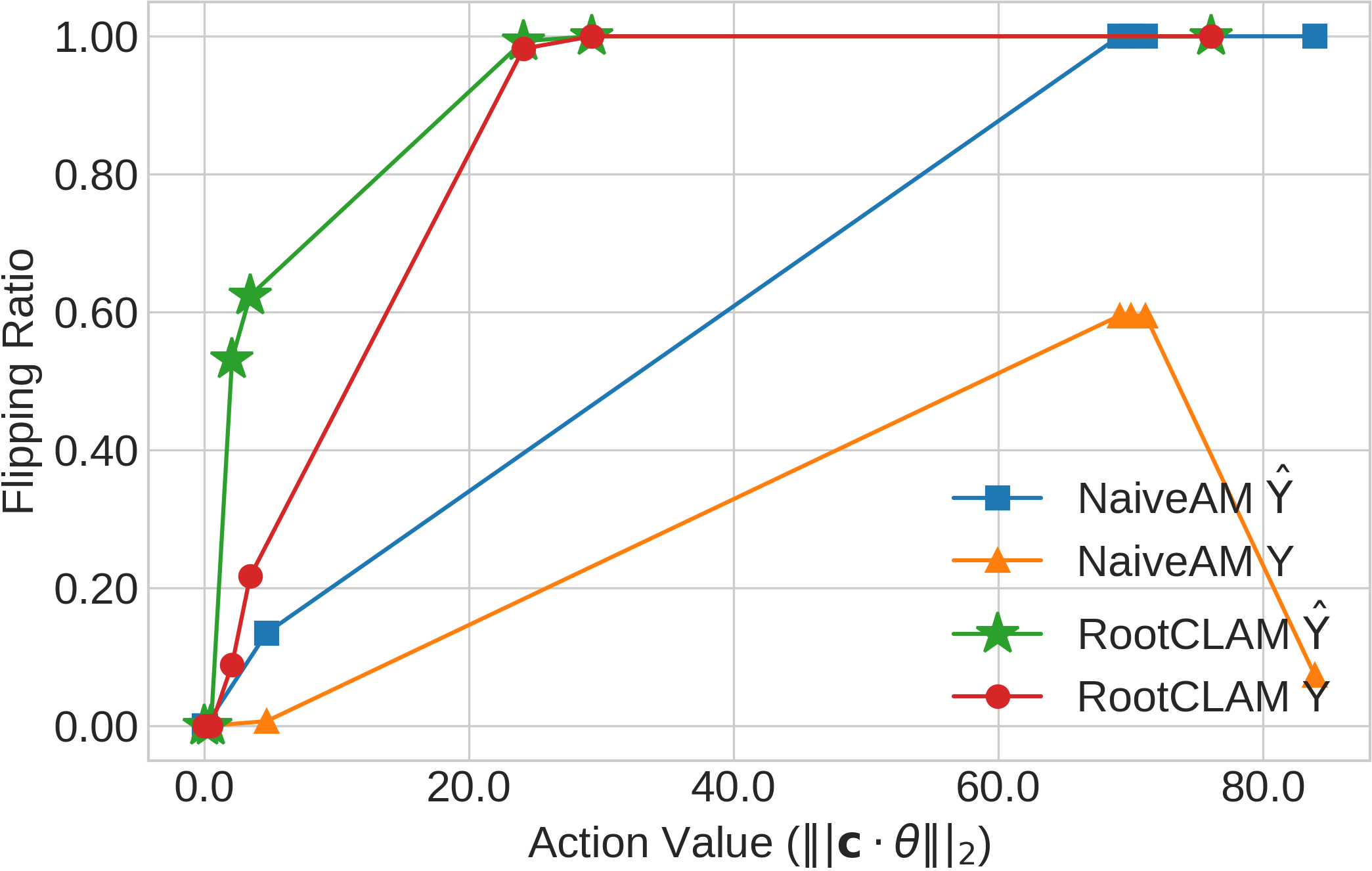}
      \caption{DeepSVDD-Adult}
      \label{fig:adult_scm}
    \end{subfigure}%
\caption{Trade-off between flipping ratio and action value.}
\label{fig:vs}
\end{figure}

{\bf \noindent The trade-off between the flipping ratio and the norm of action values.} In the objective function (Eq.~\eqref{eq:obj}), $\lambda$ as a hyper-parameter controls the trade-off between the norm of action values and the flipping ratio in the training phase. A large $\lambda$ value indicates that the model will be trained with an emphasis on minimizing the action values. Given the predicted action values, we adopt ground-truth structural equations to generate counterfactual samples and then check the flipping ratios based on anomaly detection models ($\hat{\mathbf{Y}}$) and the ground truth label $\mathbf{Y}$. Figure \ref{fig:vs} shows the results. Each point on the line from left to right indicates the result from one $\lambda$ value in the set  $[1,10^{-1},10^{-2},10^{-3},5\times10^{-4},10^{-4},10^{-5}]$. Because good mitigation action predictions should be able to flip the label with minimum changes, closing to the top-left corner indicates good performance. 

First, on both datasets, we can notice that in most cases, increasing the norm of action values can improve the flipping ratio. It means most of the abnormal samples can be flipped as normal ones with sufficient changes. Therefore, the key is to conduct minimum interventions on the original samples.  The exception is that when having a large norm of action values on NaiveAM to flip the ground-truth label $Y$, we can notice the flipping ratio either does not changes or drops, which shows the importance to consider the causal relationships when applying the mitigation actions. 

As shown in Figure \ref{fig:loan_scm_ae}, on the Loan dataset, both NaiveAM and RootCLAM can achieve a high flipping ratio evaluated by AE with very small action values ($\|\mathbf{c}\cdot \theta\|_2 < 3$). On the other hand, in terms of flipping the ground truth label $\mathbf{Y}$, RootCLAM can achieve a much higher flipping ratio compared with NaiveAM. On the Adult dataset, as shown in Figure \ref{fig:adult_scm_ae}, RootCLAM can still achieve a near 100\% flipping ratio on the detected label $\hat{\mathbf{Y}}$ as well as the ground truth label $\mathbf{Y}$, while the performance of NaiveAM is poor. 

As shown in Figure \ref{fig:loan_scm}, on the Loan dataset, both NaiveAM and RootCLAM can achieve a near 100\% flipping ratio evaluated by Deep SVDD with very small action values ($\|\mathbf{c}\cdot \theta\|_2 < 7.5$). On the other hand, in terms of flipping the ground truth label $\mathbf{Y}$, RootCLAM can achieve a higher flipping ratio with a lower norm of action values compared with NaiveAM. On the Adult dataset, as shown in Figure \ref{fig:adult_scm}, RootCLAM can still achieve better performance over NaiveAM by setting various $\lambda$ values for flipping both the ground truth label $\mathbf{Y}$ and detected label $\hat{\mathbf{Y}}$. 

\begin{figure}[t!]
\centering
    \begin{subfigure}{.23\textwidth}
      \centering
      \includegraphics[width=0.98\textwidth]{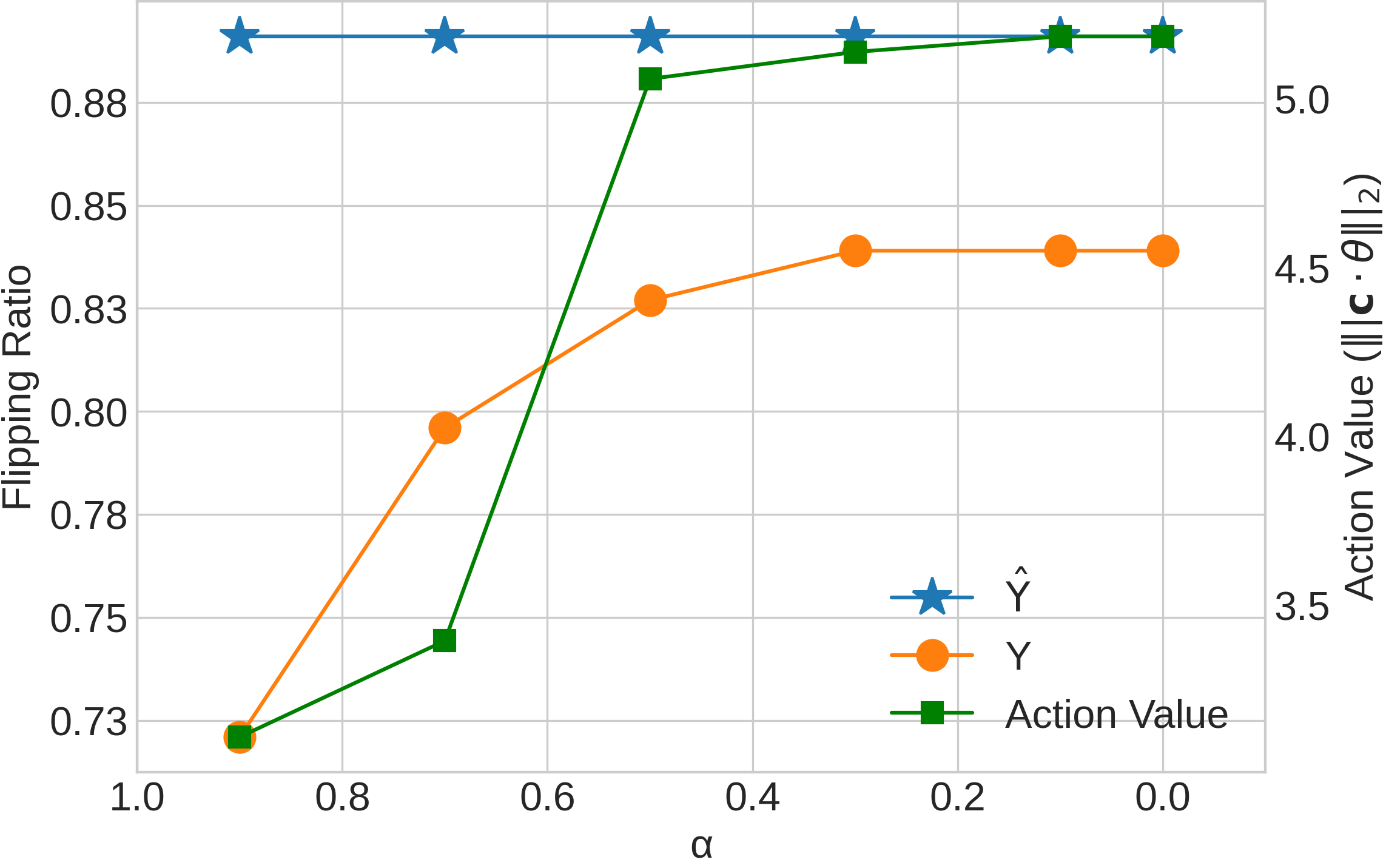}
      \caption{AE-Loan}
      \label{fig:loan_r_ae}
    \end{subfigure}%
    \hfill
    \begin{subfigure}{.23\textwidth}
      \centering
      \includegraphics[width=0.98\textwidth]{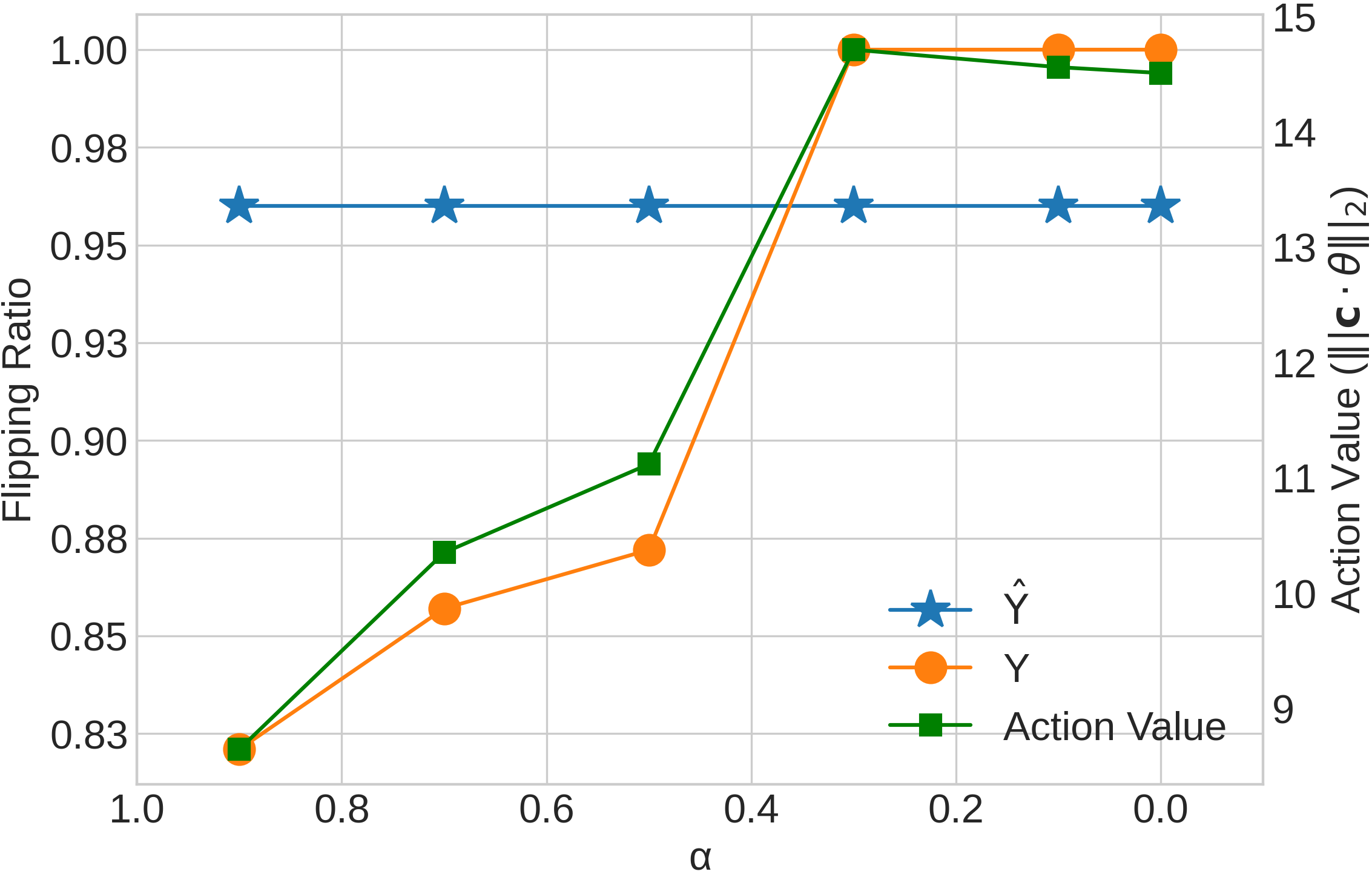}
      \caption{AE-Adult}
      \label{fig:adult_r_ae}
    \end{subfigure}%
    \hfill
    \begin{subfigure}{.23\textwidth}
      \centering
      \includegraphics[width=0.98\textwidth]{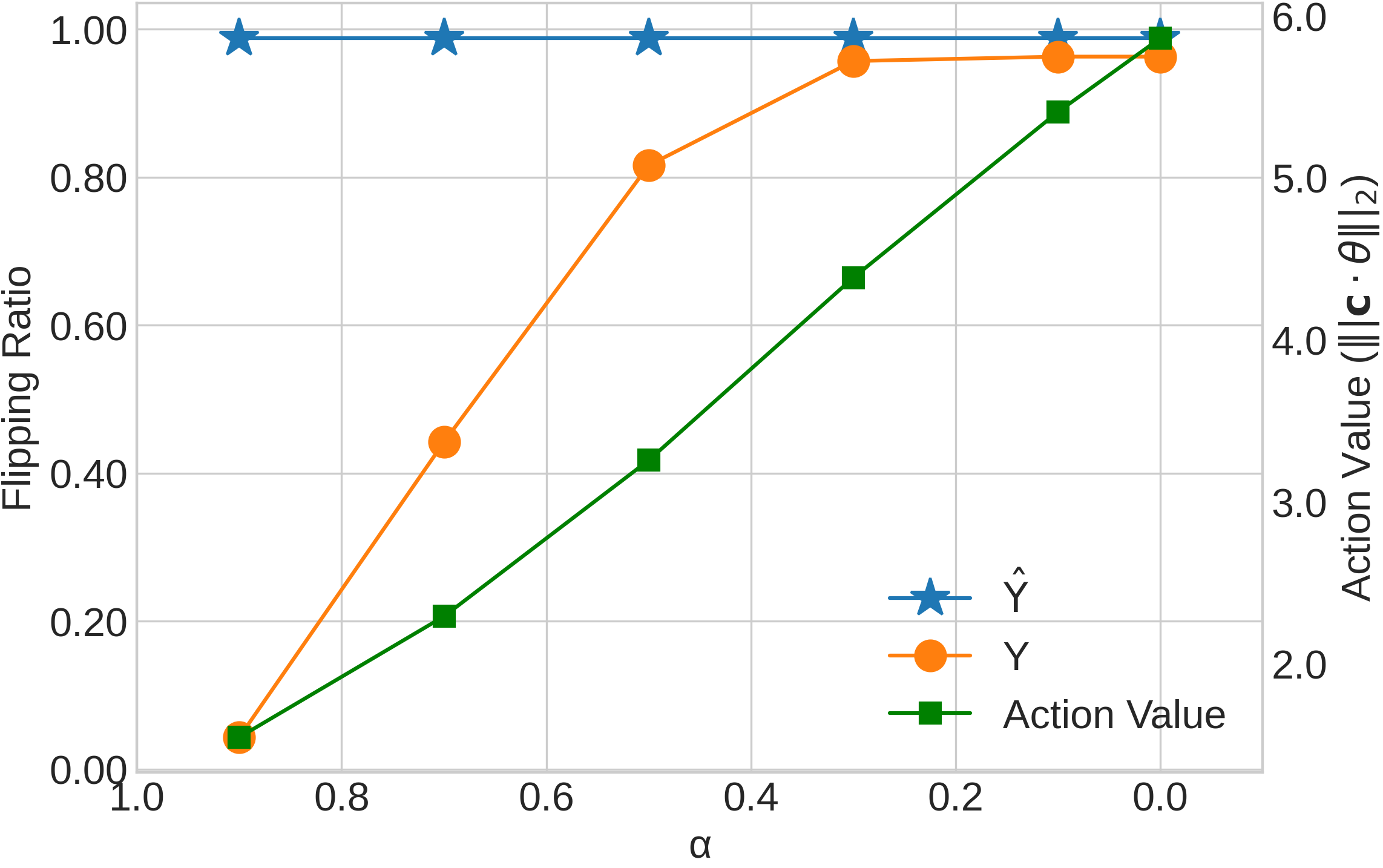}
      \caption{DeepSVDD-Loan}
      \label{fig:loan_r}
    \end{subfigure}%
    \hfill
    \begin{subfigure}{.23\textwidth}
      \centering
      \includegraphics[width=0.98\textwidth]{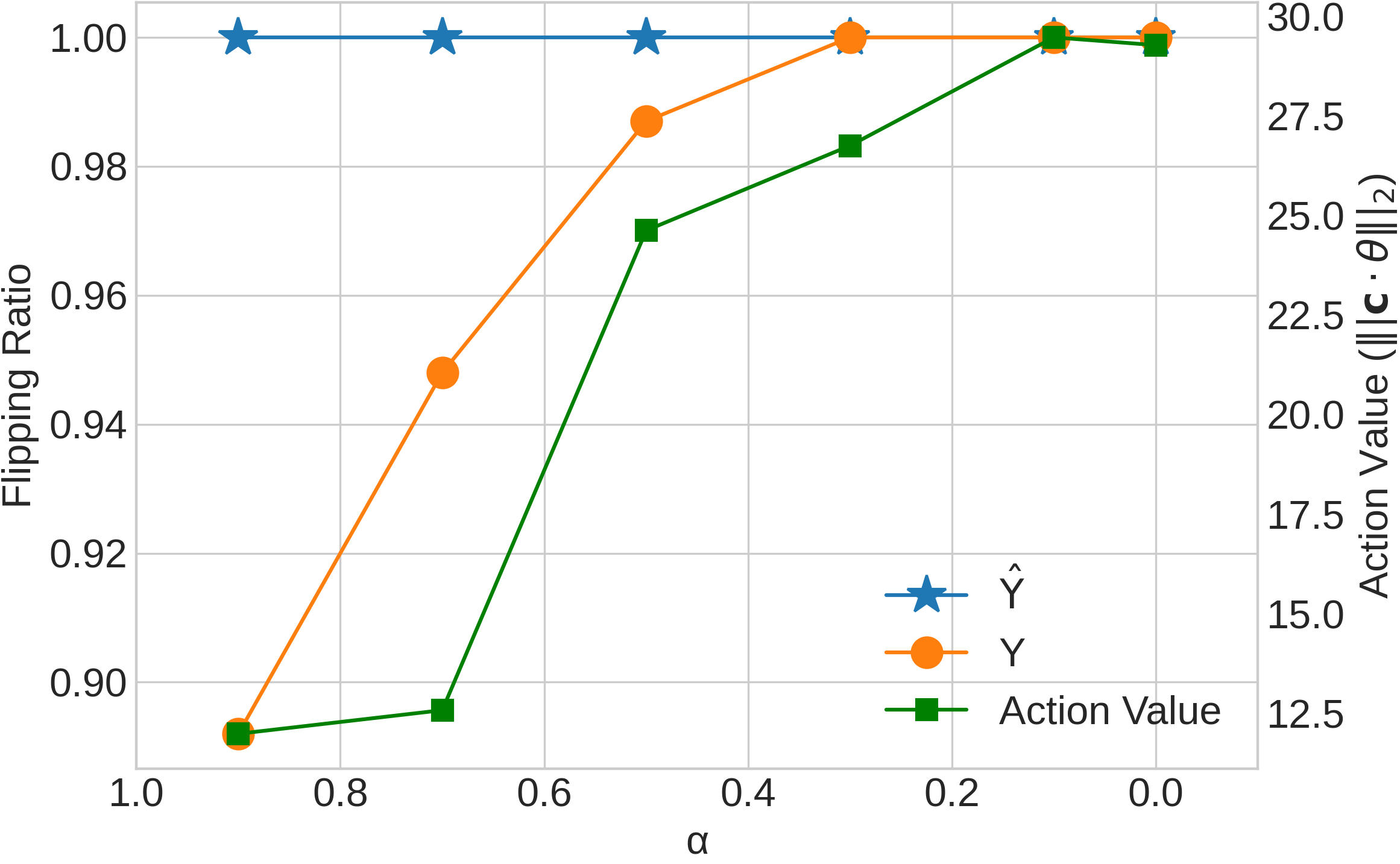}
      \caption{DeepSVDD-Adult}
      \label{fig:adult_r}
    \end{subfigure}%
   
\caption{Sensitivity analysis by setting various $\alpha$.}
\label{fig:r_res}
\end{figure}

{\bf \noindent Sensitivity analysis by setting various $\alpha$ in the objective function (Eq.~\eqref{eq:obj}) for anomaly mitigation.} The hyperparameter $\alpha$ in Eq.~\eqref{eq:obj} controls how close the anomaly scores of counterfactual samples should be to the threshold. We evaluate the flipping ratios by tuning the hyperparameter $\alpha$. 
A smaller $\alpha$ indicates that the counterfactual samples should be closer to the center of normal samples (DeepSVDD) or have a smaller reconstruction error (AE). 

Figures \ref{fig:loan_r_ae} to \ref{fig:adult_r} have similar observations. First, in all settings, the flipping ratios in terms of detected label $\hat{\mathbf{Y}}$ are high and keep stable, which shows that a small intervention on abnormal samples can flip the detecting results. Meanwhile, by reducing the $\alpha$ value, we can observe the increase of the flipping ratio in terms of ground-truth label $\mathbf{Y}$ as well as the norm of action value, which means flipping the ground-truth label requires more interventions.

\begin{figure}[t!]
\centering
    \begin{subfigure}{.23\textwidth}
      \centering
      \includegraphics[width=0.98\textwidth]{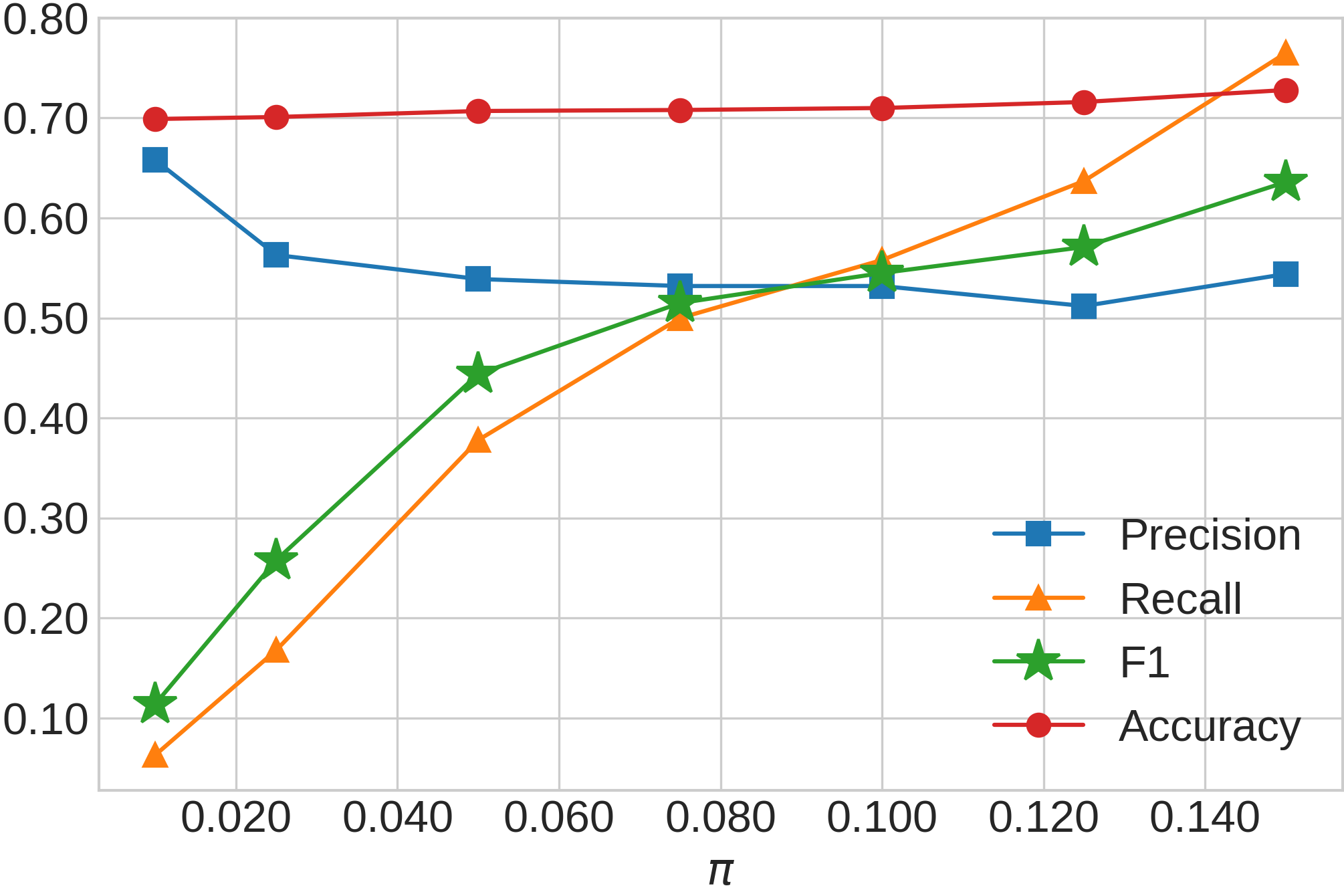}
      \caption{AE-Loan}
      \label{fig:loan_pi_ae}
    \end{subfigure}%
    \hfill
    \begin{subfigure}{.23\textwidth}
      \centering
      \includegraphics[width=0.98\textwidth]{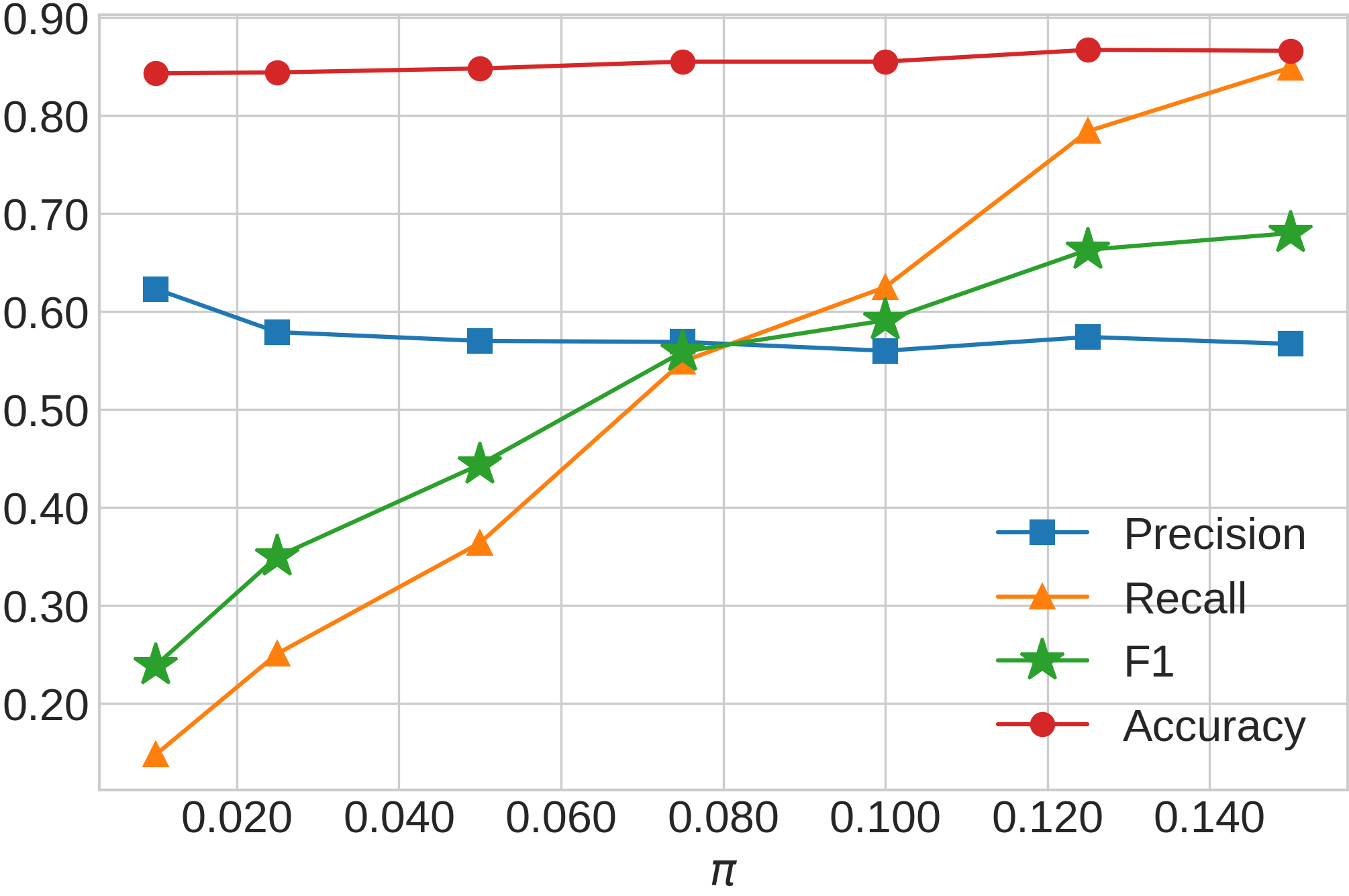}
      \caption{AE-Adult}
      \label{fig:adult_pi_ae}
    \end{subfigure}%
    \hfill
    \begin{subfigure}{.23\textwidth}
      \centering
      \includegraphics[width=0.98\textwidth]{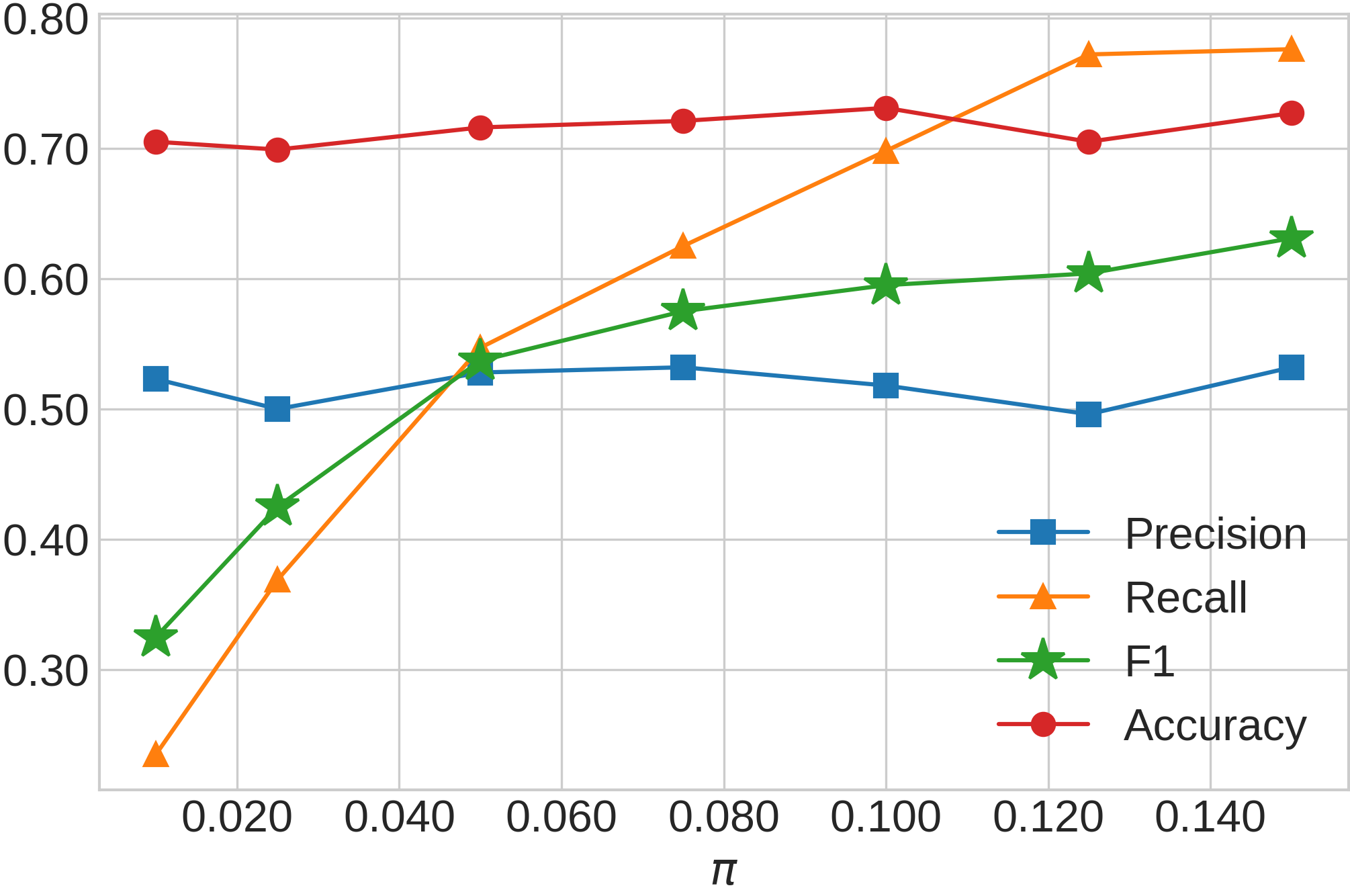}
      \caption{DeepSVDD-Loan}
      \label{fig:loan_pi}
    \end{subfigure}%
    \hfill
    \begin{subfigure}{.23\textwidth}
      \centering
      \includegraphics[width=0.98\textwidth]{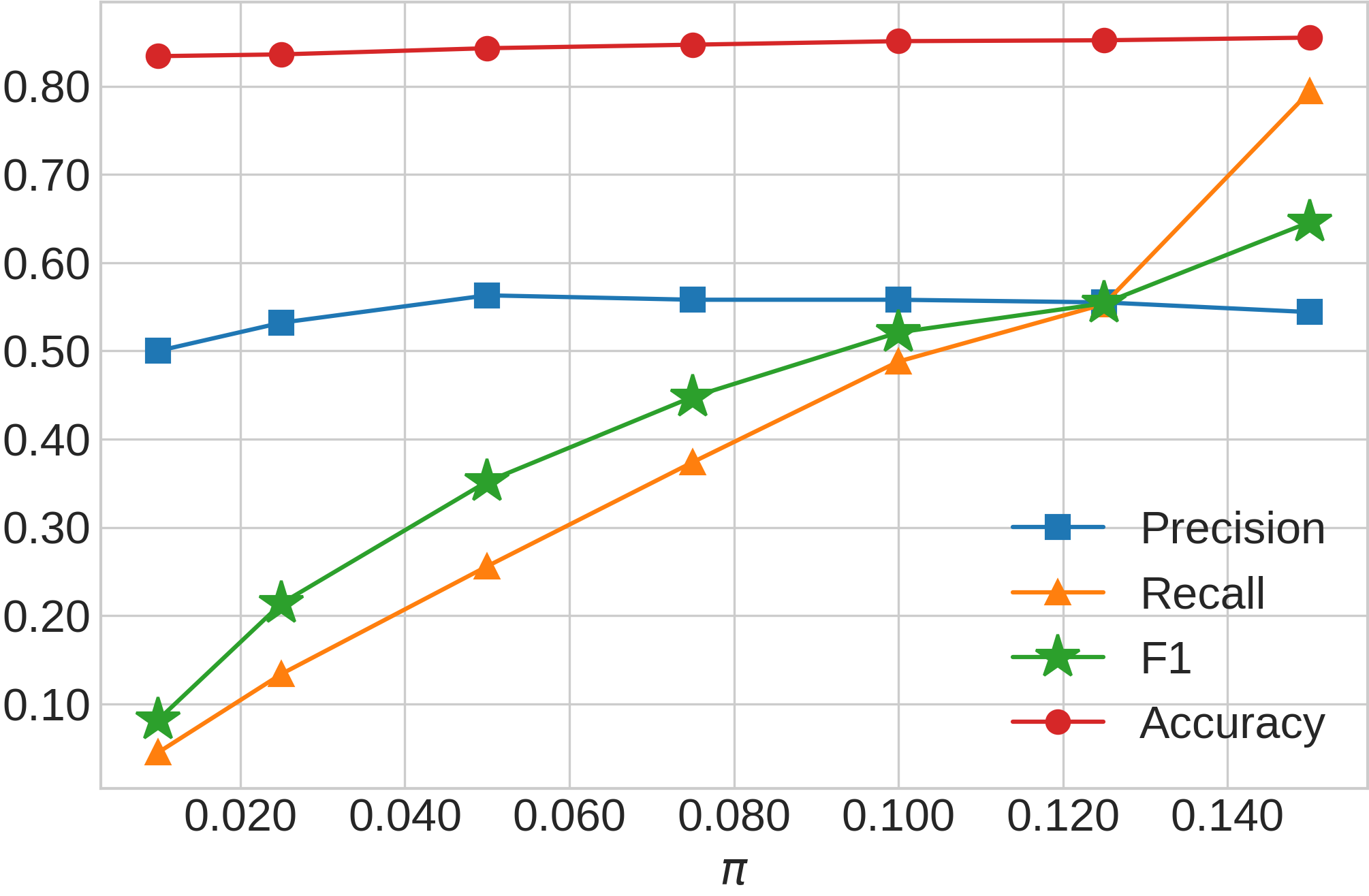}
      \caption{DeepSVDD-Adult}
      \label{fig:adult_pi}
    \end{subfigure}%
   
\caption{Sensitivity analysis by setting various $\pi$.}
\label{fig:pi_res}
\end{figure}

\begin{table*}[h]
\centering
\footnotesize
\caption{Case study on the Loan dataset, where ``loan amount'' (L) and ``loan duration'' (D) are root cause features.}
\label{tab:cs_loan}
 \begin{threeparttable}
\begin{tabular}{|c|c|c||c|c|c|c|c|c|c||c|}
\hline
\multicolumn{2}{|c|}{\multirow{2}{*}{}}                      &                                     & G & A & E & \textcolor{red}{L} & \textcolor{red}{D} & I & S & Y\\ \cline{3-11} 
\multicolumn{2}{|c|}{}                                       & $\x$                                 & 0 & -1.878 & -0.095 & 2.423 & 5.634 & -2.064 & 0.697 & 0.003\\ \hline \hline
\multirow{5}{*}{AE} &\multirow{2}{*}{NaiveAM}               & $\hat\theta$                            & / & / & / & -2.441 & -8.159 & 0.217 & -6.342 & /  \\ \cline{3-11} 
                    &                                        & $\hat{\x}(\theta)$ (SCM)              & 0 & -1.878 & -0.095 & -0.017 & -2.525 & -1.847 & -5.646 & 0.838 \\ \cline{2-11}
                    &\multirow{3}{*}{RootCLAM}                  & $\theta$                              & / & / & /  & -5.958 & -11.336 & / & / & /  \\ \cline{3-11} 
                    &                                        & $\x(\theta)$ (Eq. \ref{eq:xi})     & 0 & -2.133 & -0.089 & -3.125 & -9.154 & -1.982 & 0.162 & 0.954  \\ \cline{3-11} 
                    &                                        & $\x(\theta)$ (SCM)                    & 0 & -1.878 & -0.095 & -3.534 & -11.659 & -2.064 & 0.697 & 0.976  \\ \hline\hline
\multirow{5}{*}{Deep SVDD}  &\multirow{2}{*}{NaiveAM}               & $\hat\theta$                         & / & / & / & -1.655 & -3.911 & -1.869 & -0.161 & / \\ \cline{3-11} 
                    &                                        & $\hat{\x}(\theta)$ (SCM)              & 0 & -1.878 & -0.095 & 0.769 & 1.723 & -3.932 & 0.536 & 0.083 \\ \cline{2-11}
                    &\multirow{3}{*}{RootCLAM}                  & $\theta$                              & / & / & / & -2.157 & -12.324 & / & / & /  \\ \cline{3-11} 
                    &                                        & $\x(\theta)$ (Eq. \ref{eq:xi})     & 0 & -2.134 & -0.089 & 0.453 & -7.600 & -1.982 & 0.162 & 0.818  \\ \cline{3-11} 
                    &                                        & $\x(\theta)$ (SCM)                    & 0 & -1.878 & -0.095 & 0.267 & -8.847 & -2.064 & 0.697 & 0.850  \\ \hline
\end{tabular}

\begin{tablenotes}[para]\footnotesize
\item G -- `gender',\; A -- `age',\; E -- `education level',\; L -- `loan amount',\; D -- `loan duration',\; I -- `income',\; S -- `savings'
\end{tablenotes}
\end{threeparttable}

\end{table*}

{\bf \noindent Sensitivity analysis by setting various $\pi$ for root cause localization}.
Because the root cause features are identified with a small or large cumulative probability controlled by $\pi$, we evaluate the performance of root cause localization by tuning the threshold $\pi$. As shown in Figure \ref{fig:pi_res}, on both datasets, increasing the threshold $\pi$ can increase the recall of root cause localization with a minor negative impact on the precision. The overall performance in terms of accuracy and F1 keeps improving with a large $\pi$ value.

{\bf \noindent Case study.} 
We conduct case studies to show that RootCLAM can identify root causes and recommend mitigation actions. 
% We still use Deep SVDD as the base anomaly detection model and show the results on AE in Appendix. 
% To show the effectiveness, we also compare the actions predicted by RootCLAM with NaiveAM.

{\bf Loan Dataset.} Table \ref{tab:cs_loan} shows the case study on the Loan dataset with the root cause features $\mathcal{I}$=\{"loan amount", "loan duration"\}. For the semi-synthetic Loan dataset, the positive values of features usually indicate above the average, while negative values indicate below the average. The rows $\hat{\x}(\theta)$ (SCM) and $\x(\theta)$ (SCM) indicate counterfactual samples generated based on the structural equations given the predicted action values from NaiveAM and RootCLAM, respectively, while $\x(\theta)$ (Eq. \ref{eq:xi}) indicates the counterfactual samples generated based on our approach.

Given an abnormal sample $\x$, RootCLAM successfully identifies the two root cause features. Meanwhile, the mitigation actions predicted by RootCLAM indicate that reducing the loan amount (L) and the loan duration (D) can significantly improve the loan approval rate. 
On the other hand, although NaiveAM predicts more actions for anomaly mitigation, the odds of loan approval based on NaiveAM are still lower than the result from RootCLAM.

\begin{table*}[h]
\centering
\footnotesize
\caption{Case study on the Adult dataset, where ``hours worked per week'' (H) is the root cause feature}
\label{tb:cs_adult}

 % \resizebox{0.85\textwidth}{!}{  
 \begin{threeparttable}
\begin{tabular}{|c|c|c||c|c|c|c|c|c|c|c|c|c||c|}
\hline
\multicolumn{2}{|c|}{\multirow{2}{*}{}} & & R & A & N & S & E & \textcolor{red}{H} & W & M & O & L & I\\ \cline{3-14} 
\multicolumn{2}{|c|}{}                  & $\x$                                  & 2 & 36.401 & 1 & 1 & 5.264 & 52.520 & 1 & 1 & 2 & 1 & 60,816 \\ \hline \hline
\multirow{5}{*}{AE}                     & \multirow{2}{*}{NaiveAM}              & $\hat\theta$                          & / & 9.219 & / & / & 0.266 & -4.148 & / & / & / & / & / \\ \cline{3-14} 
                                        &                                       & $\hat{\x}(\theta)$ (SCM)              & 2 & 45.620 & 1 & 1 & 5.529 & 48.372 & 1 & 1 & 2 & 1 & 60,816 \\ \cline{2-14} 
                                        &\multirow{3}{*}{RootCLAM}                 & $\theta$                              & / & / & / & / & / & -9.672 & / & / & / & / & / \\ \cline{3-14} 
                                        &                                       & $\x(\theta)$ (Eq. \ref{eq:xi})     & 2 & 38.293 & 1 & 1 & 5.370 & 44.791 & 1 & 1 & 2 & 1 & 45,816 \\ \cline{3-14} 
                                        &                                       & $\x(\theta)$ (SCM)                    & 2 & 36.401 & 1 & 1 & 5.264 & 42.848 & 1 & 1 & 2 & 1 & 45,816 \\ \hline\hline
\multirow{5}{*}{Deep SVDD}              &\multirow{2}{*}{NaiveAM}                & $\hat\theta$                          & / & 20.734 & / & / & 0.549 & -8.573 & / & / & / & / & / \\ \cline{3-14} 
                                        &                                        & $\hat{\x}(\theta)$ (SCM)              & 2 & 57.135 & 1 & 1 & 5.813 & 43.947 & 1 & 1 & 2 & 1 & 45,816 \\ \cline{2-14} 
                                        &\multirow{3}{*}{RootCLAM}                  & $\theta$                              & / & / & / & / & / & -12.672 & / & / & / & / & / \\ \cline{3-14} 
                                        &                                        & $\x(\theta)$ (Eq. \ref{eq:xi})     & 2 & 38.293 & 1 & 1 & 5.370 & 40.217 & 1 & 1 & 2 & 1 & 45,816 \\ \cline{3-14} 
                                        &                                        & $\x(\theta)$ (SCM)                    & 2 & 36.401 & 1 & 1 & 5.264 & 39.848 & 1 & 1 & 2 & 1 & 45,816 \\ \hline                              
\end{tabular}
\begin{tablenotes}[para]\footnotesize
\item R -- `race', \; A -- `age', \; N -- `native country', \; S -- `sex', \; E -- `education level', \; H -- `hours worked per week', \; W -- `work status', \; M -- `marital status', \; O -- `occupation sector', \; L -- `relationship status', \; I -- `income'
\end{tablenotes}
 \end{threeparttable}
% }
\end{table*}

{\bf Adult Dataset.} 
Table \ref{tb:cs_adult} shows the case study on the Adult dataset with the root cause features $\mathcal{I}$=\{"hours worked per week"\}.  In this case, the action values predicted by RootCLAM on the hours worked per week is negative, which indicates that reducing hours worked per week can make the sample normal (Income less than 50k). As we consider an income higher than 50k as abnormal, our predicted action value can indicate why an individual can have a high income, i.e., having a large number of hours worked per week. On the other hand, NaiveAM cannot ensure the success of anomaly mitigation. For the AE-based model, the income value is not changed based on the action values predicted from NaiveAM. For the DeepSVDD-based model, although the action values predicted by NaiveAM successfully reduce the income, NaiveAM predicts larger action values compared to RootCLAM.

\begin{table*}[]
\footnotesize
\centering
\caption{Case study on the Donors dataset}
\label{tb:cs_donors}
% \resizebox{0.48\textwidth}{!}{  
\begin{threeparttable}
\begin{tabular}{|c|c|c||c|c|c|c|c|c|c|c|c|c||c|}
\hline
\multicolumn{2}{|c|}{\multirow{2}{*}{}}  &                                           & F1 & F2 & F3 & F4 & F5 & F6 & F7 & F8 & F9 & F10 & Y \\ \cline{3-14} 
\multicolumn{2}{|c|}{}                   & $\x$                                      & 0  & 1  & 0  & 1  & 0  & 1  & 0  & 66  & 0  & 3  & 1 \\ \hline\hline
\multirow{4}{*}{AE}         &\multirow{2}{*}{NaiveAM}   & $\hat\theta$                               & /  & /  & /  & /  & /  & /  & /  & 34  & 5 & -5 & /\\ \cline{3-14}
                            &                           & $\hat{\x}(\theta)$ (Eq. \ref{eq:baseline})    & 0  & 1  & 0  & 1  & 0  & 1  & 0  & 100  & 5 & -3 & 1 \\ \cline{2-14}
                            &\multirow{2}{*}{RootCLAM}      & $\theta$                                  & /  & /  & /  & /  & /  & /  & /  & 26 & 2  & 5   & /\\ \cline{3-14}
                            &                           & $\x(\theta)$ (Eq. \ref{eq:xi})         & 1  & 1  & 1  & 1  & 1  & 1  & 0  & 100 & 2  & 6  & 0 \\ \hline\hline
\multirow{4}{*}{Deep SVDD}  &\multirow{2}{*}{NaiveAM}   & $\hat\theta$                              & /  & /  & /  & /  & /  & /  & /  & 34  & 5 & -4 & /\\ \cline{3-14}
                            &                            & $\hat{\x}(\theta)$ (Eq. \ref{eq:baseline})        & 0  & 1  & 0  & 1  & 0  & 1  & 0  & 100  & 5 & -2 & 1 \\ \cline{2-14}
                            &\multirow{2}{*}{RootCLAM}      & $\theta$                                  & /  & /  & /  & /  & /  & /  & /  & 26 & 2  & 5   & /\\ \cline{3-14}
                            &                            & $\x(\theta)$ (Eq. \ref{eq:xi})         & 1  & 1  & 1  & 1  & 1  & 1  & 0  & 100 & 2  & 6  & 0 \\ \hline
\end{tabular}
\begin{tablenotes}[para]\footnotesize
\item F1 -- `at least 1 teacher-referred donor', \; F2 -- `fully funded', \; F3 -- `at least 1 green donation', \; F4 -- `great chat', \; F5 -- `three or more non teacher-referred donors', \; F6 -- `one non teacher-referred donor giving 100 plus', \; F7 -- `donation from thoughtful donor', \; F8 -- `great messages proportion', \; F9 -- `teacher-referred count', \; F10 -- `non teacher-referred count'. 
\end{tablenotes}
\end{threeparttable}
% }
\end{table*}                                 

{\bf Donors Dataset.} We consider a project that is not exciting as an anomaly and aim to flip the label. Based on the definition of an exciting project, the original sample $\x$ in Table \ref{tb:cs_donors} is not exciting because this project fails to meet the requirements of at least one teacher-referred donor (F1) and at least one ``green'' donation (F3). In this case study, RootCLAM identifies ``great messages proportion'' (F8), ``teacher-referred count'' (F9), and ``non teacher-referred count'' (F10) as the root cause features. All root cause features are ancestors of exciting requirements shown in Figure \ref{fig:graph_donors}. After getting the action values from $h_\phi(\cdot)$, we round to the nearest integer. Because we do not have the ground truth structural equations for Donors, Table \ref{tb:cs_donors} only shows the predicted counterfactual samples from the models.

For the purpose of anomaly mitigation, in order to make the project exciting, as shown in Table \ref{tb:cs_donors}, the project host should try to have more `great messages', increase the `teacher-referred count' as well as `non-teacher-referred count'. After doing such changes, as shown in the last row, some key features, such as F1, F3, and F5, are flipped to 1.  Then, we can notice that the counterfactual sample will be exciting. On the other hand, because NaiveAM does not consider the causal relationships among features, NaiveAM cannot derive the impact on other features after changing the root cause features. 
As a result, NaiveAM cannot flip the label. 

\section{Conclusion}
In this paper, we developed RootCLAM, a framework for root cause analysis and anomaly mitigation through causal inference. RootCLAM first learns a Variational Causal Graph Autoencoder from the normal data. Then, given an abnormal sample, RootCLAM identifies root cause features with the exogenous variables significantly deviated from the regular data. Then, RootCLAM computes mitigation actions as soft interventions on root cause features that can flip the anomalies to normal. Experiments show that RootCLAM achieves state-of-the-art performance on root cause localization and can further successfully fix most of the anomalies. 

\section*{Acknowledgement}
This work was supported in part by NSF 1910284 and 2103829.

\clearpage
\bibliographystyle{ACM-Reference-Format}
\balance
\bibliography{main}

\end{document}